\documentclass{osa-article}

\journal{oe}
\articletype{Research Article}

\usepackage{color}
\newtheorem{myDef}{Definition}

\begin{document}
	
\title{Matching entropy based disparity estimation from light field}
	
\author{Ligen Shi,\authormark{1} Chang Liu,\authormark{2} Di He,\authormark{2} Xing Zhao,\authormark{1,*} and Jun Qiu\authormark{2,*}}
	
\address{\authormark{1}School of Mathematical Science, Capital Normal University, Beijing 100048, China\\
\authormark{2}Institute of Applied Mathematics, Beijing Information Science and Technology University, Beijing 100101, China}
\email{ \authormark{1,*}zhaoxing\_1999@126.com\\
	\authormark{2,*}qiu.jun.cn@ieee.org}
 
\begin{abstract}
A major challenge for matching-based disparity estimation from light field data is to prevent mismatches in occlusion and smooth regions. An effective matching window satisfying three characteristics: texture richness, disparity consistency, and anti-occlusion should be able to prevent mismatches to some extent. According to these characteristics, we propose matching entropy in the spatial domain of the light field to measure the amount of correct information in a matching window, which provides the criterion for matching window selection. Based on matching entropy regularization, we establish an optimization model for disparity estimation with a matching cost fidelity term. To find the optimum, we propose a two-step adaptive matching algorithm. First, the region type is adaptively determined to identify occluding, occluded, smooth, and textured regions. Then, the matching entropy criterion is used to adaptively select the size and shape of matching windows, as well as the visible viewpoints. The two-step process can reduce mismatches and redundant calculations by selecting effective matching windows. The experimental results on synthetic and real data show that the proposed method can effectively improve the accuracy of disparity estimation in occlusion and smooth regions and has strong robustness for different noise levels. Therefore, high-precision disparity estimation from 4D light field data is achieved.
\end{abstract}

\section{Introduction}
A light field\cite{levoy1996light,1995Plenoptic} record the spatial and angular information of a set of light rays in the scene space and have been widely used in scene depth estimation and three-dimensional (3D) imaging\cite{2001Integral,2013Advances, MartinezCorral18}. From the perspective of data acquisition, the light field data can be acquired directly by imaging devices, and indirectly reconstructed from focal stacks or encoded masks. Integral imaging and camera arrays are two basic types of direct acquisition imaging systems. Gabriel Lippmann proposed integral photography in 1908 and captured the spatial-angular information of 3D scenes for the first time\cite{lippmann1908epreuves,sokolov1911autostereoscopy,ives1931optical}. Integral imaging\cite{lippmann1908epreuves,martinez2018fundamentals,martinez2017recent,javidi2020roadmap} can be regarded as a 3D imaging technique that captures and reproduces a light field by using a two-dimensional (2D) array of microlenses (or lenslets).  In light field capture mode, in which the detector is coupled to the microlens array, each microlens allows an image of the subject as seen from the viewpoint of that lens's location to be acquired. The display manner using integral imaging can be regarded as a type of light field display. In the reproduction mode, in which an object or source array is coupled to the microlens array, each microlens allows each observing eye to see only the area of the associated micro-image containing the portion of the subject.

In terms of theoretical modeling, E. H. Adelson et al. proposed the seven-dimensional (7D) plenoptic function $L(V_x, V_y, V_z,\phi,\varphi,\lambda,t)$ to describe the irradiance information of a light ray with any wavelength in space at any time\cite{Adelson1991plenoptic}. Then, a 4D light field, which is a simplified two-plane representation suitable for optical imaging systems, was developed in many integral imaging systems\cite{Ng2005LightFP} and camera array systems.  The integral imaging systems' optical geometry can be implemented and visualized by substituting pinholes for the micro-lenses, as has been done for some demonstrations and special applications.  The fundamentals, related techniques, and emerging applications of light field data and integral imaging techniques for 3D imaging and displays have been extensively studied and comprehensively summarized \cite{martinez2018fundamentals,martinez2017recent,javidi2020roadmap}.

 Scene disparity estimation from light field data is an essential problem in light field computational imaging, especially in 3D digital imaging. There are four categories of methods used to estimate disparity information from a 4D light field: matching-based, epipolar-geometry-based, focus-measure-based, and deep-learning-based methods. Matching-based methods\cite{2016Reliable,2015Accurate,2014Light,2020Fast,2014Shape,2013Variational,2013Stereo} is an extension of the stereo matching method, which can reduce the influence of light field spectrum aliasing and angular artifacts. However, matching often fails in smooth and occlusion regions. Epipolar plane images (EPIs) reveal the epipolar geometry of light fields\cite{2013Globally,2015Continuous,2016What,6247656}. Therefore, the depth can be obtained by calculating the slope of the epipolar line in an EPI. Epipolar-geometry-based methods can achieve good results in occlusion regions, but they require a large amount of calculation and are sensitive to noise. Focus-measure-based methods obtain the depth by the focus measure in the focal stack\cite{ Tao13,2016Robust,2016Occlusion,7987707}. Since the focal stack is the projection of the light field in the preset depth layers, the estimation accuracy depends on the sampling of depth layers. Deep-learning-based methods replace complex depth estimation pipelines with neural networks\cite{Jeon2019Depth, Shin_2018_CVPR,7780983,2020Attention}, which require a large amount of training data and lack generalization abilities.

Area matching is a commonly used technique in matching-based methods and makes use of window matching instead of pixel matching to improve the robustness. A unified matching window leads to calculation redundancy in textured regions and mismatches in occlusion and smooth regions. If we can determine the occluding, occluded, smooth, and textured regions, we will be able to improve the estimation accuracy and efficiency by selecting the effective window size and shape for different regions. For textured and smooth regions, selecting a matching window that covers enough textures is the key task. For occluding regions, selecting the shapes according to the occlusion geometry is the key task. For occluded regions, selecting the shape of the matching windows and the visible viewpoints are the key tasks.

To accomplish the key tasks, an effective matching window should satisfy three characteristics: texture richness, disparity consistency, and anti-occlusion, thus will provide enough valid matching information and less invalid or incorrect information. We propose matching entropy corresponding to these characteristics to measure the effectiveness of a matching window. With matching entropy acting as the regularization term, we establish an optimization model for disparity estimation and propose a two-step adaptive window matching method to solve the optimization model. In the first step, the region type is adaptively determined based on the segmentation and the texture information. In the second step, matching entropy is used as a criterion for the adaptive selection of the matching windows’ shape and size, and the visible viewpoints.
\begin{figure}[ht!]
		\centering\includegraphics[width=13cm]{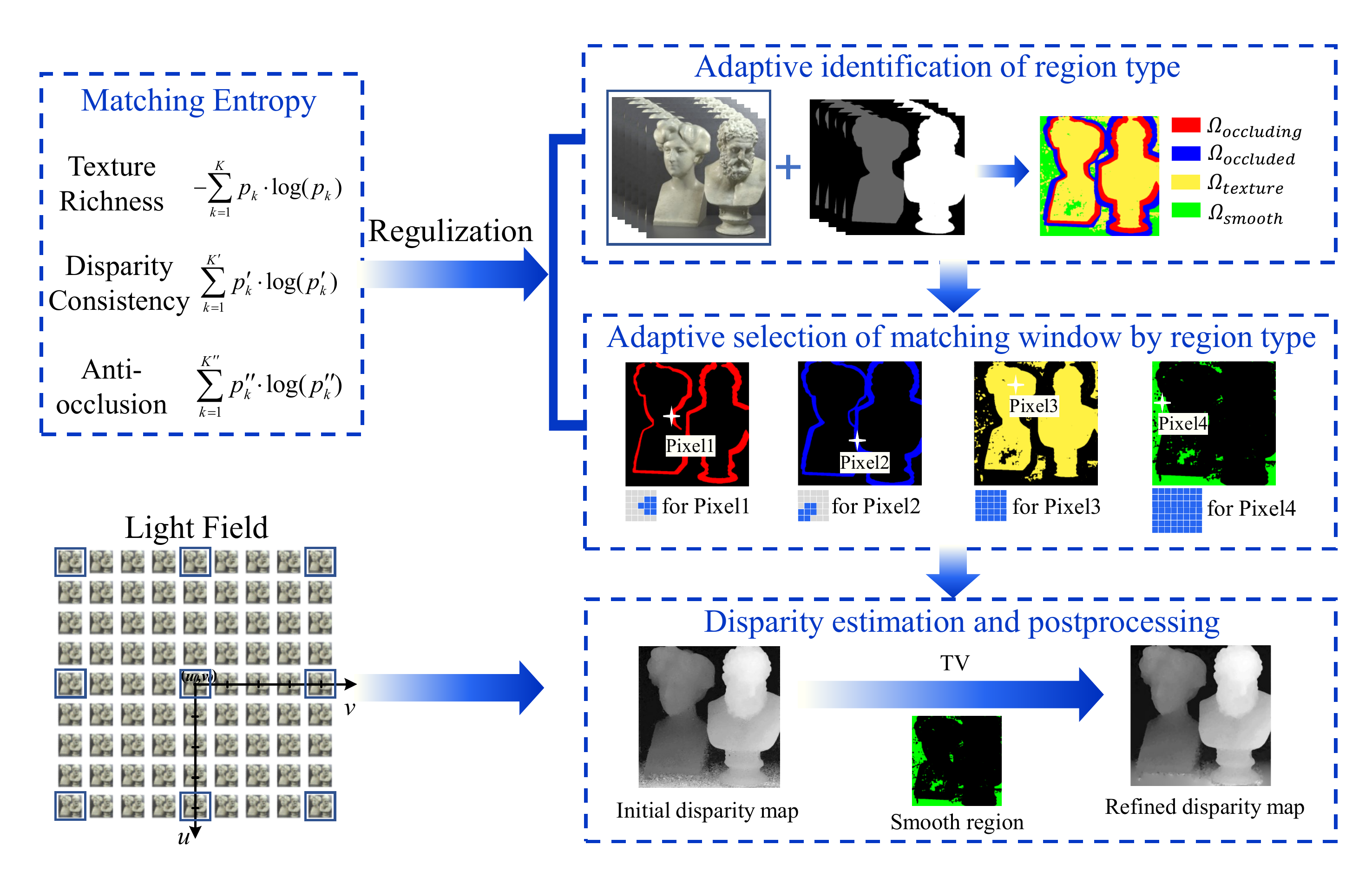}
\caption{Scheme of the proposed matching entropy based disparity estimation.}
		\label{fig1}
\end{figure}

\section{Related work}
 The main implementation of light field computational imaging and computational display is integral imaging. The resolution and field of view of light field data depend on the capability of integral imaging systems. The optimum lenslet size in the lenticular screen and the resolution limitation for integral imaging were derived \cite{burckhardt1968optimum}. To improve the real-time performance of integral imaging systems, a real-time integral imaging method\cite{arai1998gradient} was proposed to provide 3D autostereoscopic images of moving objects in real time by using microlens arrays. B. Javidi et al. proposed synthetic aperture integral imaging\cite{jang2002three}, in which an effectively extended FOV is obtained by moving a small integral imaging system, which greatly increases the FOV and resolution. The synchronously moving micro-optics (lenslet arrays) was utilized in an integral imaging system for image capture and display, in order to overcome the resolution limitation by the Nyquist sampling\cite{jang2002improved}. F. Jin et al. clarified the effects of a finite number of pixels in elemental images on the resolution and the depth of focus in three-dimensional integral imaging\cite{jin2004effects}.

 Since the light field data is the sensing, visualization and 3D display of the scene information, integral imaging systems are 
practical for many fields. S.H. Hong et al. proposed a 3D imaging technique based on integral imaging \cite{hong2004three}, which can perceive 3D scenes and reconstruct them into 3D volumetric images. The reconstruction of scene volume pixels is implemented by simulating optical reconstruction based on ray optics calculations. H. Arimoto et al. reconstructed 3D images by numerically processing an array of observed images formed by a microlens array\cite{arimoto2001integral}. The algorithms for reconstructing 3D images are robust and can obtain images viewed from arbitrary directions. A. Stern et al. proposed a computational synthetic aperture integral imaging technique \cite{stern20033}, which can increase the field of view (FOV). The synthetic aperture is obtained by the relative motion of the imaging system and the object in a plane perpendicular to the optical axis. C.G. Luo et al. analyzed the depth of field (DOF) of integral imaging display based on wave optics\cite{luo2013analysis}. Considering the diffraction effect, the intensity distribution of light with multiple microlenses is analyzed, and the formula for calculating the DOF of the integral imaging display system is derived. 

 As a middle-level vision process in light field imaging, disparity estimation is an essential topic for high-precision 3D visual perception and high-fidelity 3D content generation. The applications of light field imaging, such as light field super-resolution, digital refocusing, light field compression, and light field editing, largely depend on the accurate estimation of disparity (or depth). In recent years, researchers in the field of optics have also been focusing on the subject of disparity estimation from light field data. The disparity resolution properties of light field data were analyzed \cite{ma2017depth} in case of limiting the epipolar analysis to a small range to reduce runtime, combined with regression testing to reduce estimation error. The iterative scheme was proposed for fidelity reconstruction of scene depth from 4D light field data\cite{liu2017iterative}. A novel active disparity estimation method\cite{cai2019accurate} was proposed by directly using the corresponding cues in structured light fields to search for the unambiguous disparity. The geometric model based on epipolar space\cite{zhou2019light}, was proposed to determine the relationship between 3D points in a scene and the 4D light field, then the closed-form solution for geometric-model-based 3D shape measurement is completed. The influence of plenoptic imaging distortion on light field disparity estimation\cite{cai2020light} was clarified to propose the light field disparity estimation method considering plenoptic imaging distortion. In addition, an accuracy analysis of light field depth estimation is performed using standard phantoms. To handle different types of occlusion, S. Ma et al. proposed the side window subsets for angular coherent\cite{ma2021occlusion} and theoretically analyzed the ability of the proposed method to resist occlusions. Deep learning methods are explored for predicting scene disparity. X. Wang et al. proposed a convolutional neural network based on epipolar geometry and image segmentation for light field disparity estimation\cite{wang2020light}. Multi-directional epipolar images are chosen as input data, and convolutional blocks are employed according to the disparity of different directional epipolar images. B. Liu et al. proposed a light field disparity estimation network\cite{liu2022cascade}, which employs a cascaded cost volume architecture that can predict disparity maps in a coarse-to-fine manner by fully exploring the geometric features of sub-aperture images. 

 The scene disparity estimation approach that matches sub-aperture image arrays comes from area matching in stereo matching\cite{2002Multi}, because the sub-aperture images can be regarded as multiview images. The designs of matching windows and the matching costs are the key problems of area matching. Typical matching windows include weighted windows \cite{2014Hardware}, reliable multi-scale and multi-windows (MSMWs)\cite{2015Reliable}, and cross-based local windows\cite{4811952}. A weighted window is a fixed shape window with radiance- or distance-based weights for pixels. An MSMW is selected from a window dictionary by minimizing the matching cost. A cross-based window is generated by a crisscross expansion of the anchor pixel according to the color consistency. The commonly used matching costs include the sum of absolute differences (SAD), the sum of squared differences (SSD), the normalized cross-correlation (NCC), and census\cite{10.1007/BFb0028345}.

Sub-aperture images can be regarded as a dense uniform sampling of the viewpoint plane with a small baseline. The small baseline leads to subpixel disparities, which can hardly be detected using conventional matching methods. Spatial interpolation can be used to solve this problem to a certain extent. However, the blur caused by the interpolation increases the possibility of mismatches. H. G. Jeon et al.\cite{2015Accurate} proposed the phase shift theorem that allows the estimation of the subpixel offset between sub-aperture images. To reduce the mismatches in the occlusion region, J. Navarro et al.\cite{2016Reliable} used an MSMW\cite{2015Reliable} to estimate the disparity between the central view and the rest of the views in the same row and column and then used the median operator to extract the reliable disparity value. C. Chen et al.\cite{2014Light} proposed a bilateral metric considering the color consistency and the pixel distance in the reference window to improve the robustness in occlusion regions, but this method is sensitive to noise. W. Williem et al.\cite{2016Robust} proposed the analysis of angular patches to form a matching cost by combining the angular entropy metric and adaptive defocus response. The angular entropy metric is more robust to occlusion but sensitive to noise. The balance between angular entropy and the adaptive defocus response is intractable. T. C. Wang et al.\cite{2016Occlusion} proposed an occlusion-aware disparity estimation cross-correlation with occluded line edges. The accuracy of the disparity estimation result is highly dependent on edge detection. Using occlusion-noise-aware data costs, the constrained entropy cost in the angular domain of the light field is proposed to reduce the effects of the dominant occluder and noise in the angular patch, resulting in a low cost\cite{park2017robust}. For super-resolution and disparity estimation, a generic mechanism to disentangle the coupled information for LF image processing, and a class of domain-specific convolutions is designed to disentangle LFs from different dimensions\cite{wang2022disentangling}.

To reduce the mismatches in occluded and smooth regions, we propose matching entropy in the spatial domain of the light field to measure how well a matching window in different regions meets the three characteristics. The optimization model based on matching entropy regularization is utilized for disparity estimation in occlusion, smooth, and textured regions.
	
\section{The optimization model based on matching entropy regularization}
The fixed window for region matching may lead to mismatches between occlusion regions and smooth regions. An effective way to solve the mismatch is to eliminate the mismatched part in the window that generates the mismatch and increase the amount of the information that could match correctly. In this paper, the shape of the matching window is used to eliminate the mismatched part, and the size of the matching window is applied to increase the amount of information that can correctly match the region. We proposed the matching entropy measure the correct information in a matching window, and hence, it becomes a criterion for the matching window selection. 
\subsection{Matching entropy}
To estimate the depth map accurately, every matching window $w(x,y)$ needs to contain a sufficient amount of effective matching information. The ideal matching window should satisfy three characteristics: texture richness, disparity consistency, and anti-occlusion. Texture richness is fundamental for area matching. Disparity consistency is the basic assumption of area matching, which ensures that the area remains invariant in different view images. Anti-occlusion is essential for accurate and robust matching in occlusion regions. According to these characteristics, we define the matching entropy of a window $w(x,y)$ to measure the amount of effective matching information.
	
\begin{myDef}
For a light field $L(u,v,x,y)$, the matching entropy of a window $w(x,y)$ in the central view image $L_{\left(u_{0}, v_{0}\right)}(x, y)$ is defined as
\begin{equation}
		E^{entropy}[w(x, y)]=-\sum_{k=1}^{K} p_{k} \cdot \log \left(p_{k}\right)+\alpha_{1} \sum_{k=1}^{K^{\prime}} p_{k}^{\prime} \cdot \log \left(p_{k}^{\prime}\right)+\alpha_{2} \sum_{k=1}^{K^{\prime \prime}} p_{k}^{\prime \prime} \cdot \log \left(p_{k}^{\prime \prime}\right),
\label{eq1}
\end{equation}
where $p_k$ and $p_k^{\prime}$ stand for the probabilities of the gray value and the disparity value of the $k\-th$ pixel in $w(x,y)$, respectively, while ${p}_{k}^{\prime\prime}$ is the probability of the gray value of the $k\-th$ pixel of the mismatched pixels in $w(x,y)$. ${\alpha}_{1}\geq{0}$ and  ${\alpha}_{2}\geq{0}$ denote the weight coefficients. ${\alpha}_\mathbf{2}={0}$ when there is no occlusion in $w(x,y)$.
\end{myDef}

The three terms of the matching entropy function refer to texture richness, disparity consistency, and anti-occlusion respectively. $p_k$ and $p_k^{\prime}$ are calculated from the gray histogram and the disparity histogram of $w(x,y)$, respectively, and ${p}_{k}^{\prime\prime}$ is obtained from the gray histogram of the mismatched pixels in $w(x,y)$. In the anti-occlusion term, the mismatched pixels are the occluded pixels in $w(x,y)$ if Pixel $(x,y)$ occludes other pixels, and are the occluding pixels if Pixel $(x,y)$ is occluded.
	
\subsection{The optimization model}

\begin{figure}[ht!]
\centering\includegraphics[width=13cm]{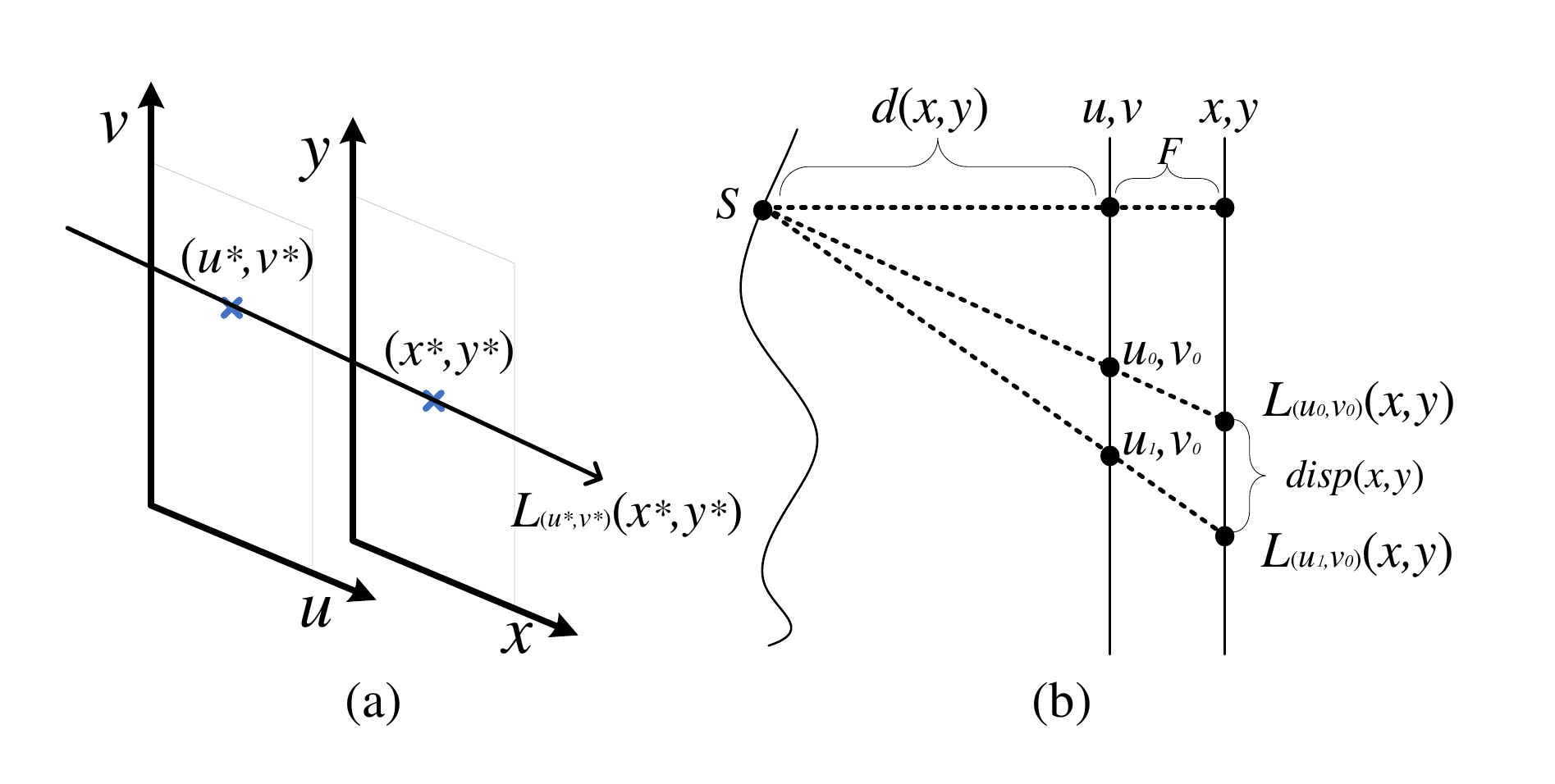}
\caption{(a) Two-plane parameterized light field. (b) Diagram of the scene point projected in different views. $(u,v)$ and $(x,y)$ represent the viewpoint plane and the image plane, respectively. $F$ is the distance between the two parameterization planes, and $(u_0, v_0)$ the central viewpoint.  $d(x,y)$ and $disp(x,y)$ are the depth and the disparity of the scene point ${S}$, 
 respectively.}
\label{fig2}
\end{figure}

In the light field data, the scene point is projected in different views\cite{wanner2013variational,sheng2018occlusion,liu2017iterative} as shown in Fig.~\ref{fig2}. Under the perspective projection, the scene points projected onto the $(x,y)$ plane form a scene surface in 3D space. The depth $d(x,y)$, disparity $disp(x,y)$, and scene surface $\vec{S}(x,y)$ are all represented and defined in the same coordinate system $(x,y)$.

The relationship between a scene point $\vec{S}(x,y)$  and its depth map $d(x,y)$ is

\begin{equation}
		\vec{S}(x, y)=\left(-\frac{F x}{d(x, y)}+u_{0},-\frac{F y}{d(x, y)}+v_{0}, d(x, y)\right)^{T}
\label{eq2}.
\end{equation}

The relationship between a scene point $\vec{{S}}\left({x},{y}\right)$ and its disparity ${disp}\left({x},{y}\right)$ is
\begin{equation}
		\vec{S}(x, y)=\left(-x({disp}(x, y)-1)+u_{0},-y({disp}(x, y)-1)+v_{0}, \frac{F}{d i s p(x, y)-1}\right)^{T}
		\label{eq3}.
	\end{equation}

The coordinates of $\vec{{S}}\left({x},{y}\right)$ from viewpoint $({u},{v}) $ are denoted as $\left({x},{y}\right)_{{u},{v}}$. Then, the relationship between the image coordinates under viewpoints $\left({u}_\mathbf{0},{v}_\mathbf{0}\right)$ and $({u},{v})$ is
\begin{equation}
		(x, y)_{u, v}=(x, y)_{u_{0}, v_{0}}+\left(\left(u-u_{0}\right) \cdot {disp}(x, y),\left(v-v_{0}\right) \cdot {disp}(x, y)\right)
		\label{eq4}.
\end{equation}

According to Eq.~\ref{eq4}, the matching term acting as the fidelity term of the optimization model can be defined as 
\begin{equation}
		\begin{array}{l}
			E^{\operatorname{match}}[disp(x, y), w(x, y)]\\
			=\sum\limits_{(u, v) \in \Phi} \sum\limits_{(m, n) \in w(x, y)} \beta_{(m, n)}\left\|L_{u_{0}, v_{0}}\left((x, y)_{u_{0}, v_{0}}+(m, n)\right)-L_{u, v}\left((x, y)_{u, v}+(m, n)\right)\right\| \\
			=\sum\limits_{(u, v) \in \Phi} \sum\limits_{(m, n) \in w(x, y)} \beta_{(m, n)}\|L_{u_{0} v_{0}}\left((x, y)_{u_{0}, v_{0}}+(m, n)\right)-L_{u, v}((x, y)_{u_{0}, v_{0}}\\
			+\left(\left(u-u_{0}\right) \cdot disp(x, y),\left(v-v_{0}\right) \cdot disp(x, y)\right)+(m, n))\|	
			\label{eq5},
		\end{array}
	\end{equation}
where ${\Phi}$ represents the set of visible viewpoints in the matching window ${w}\left({x},{y}\right)$, ${\beta}_{(m, n)}=1/\#\left\{w(x,y) \right\}$ is the weight coefficient, and $\#\left\{\cdot \right\}$ denotes the number of pixels in the window. 

Combining the matching entropy, we established the objective functional for disparity estimation.
\begin{equation}
		E[disp(x, y), w(x, y)]=E^{\text{match }}[disp(x, y), w(x, y)]-\lambda E^{\text {entropy}}[w(x, y)]
		\label{eq6},
	\end{equation}
where ${E}^{\text{entropy}}[w(x,y)]$ is the matching entropy term acting as the regularization term and ${\lambda}$ is the regularization parameter.

By solving the following optimization problem, the effective matching windows can be selected in light of matching entropy, and meanwhile the disparity map can be estimated.
\begin{equation}
		\left[disp^{*}(x, y), w^{*}(x, y)\right]=\arg\min_{disp, w}(E(disp(x, y), w(x, y)))
		\label{eq7}.
	\end{equation}

Since the optimal matching windows contain sufficient effective information without mismatching information, minimizing the fidelity term with the optimal windows can realize accurate and robust disparity estimation.
	
\section{Implementation of disparity estimation by adaptive region matching}
\subsection{Adaptive identification of region types}
Since the selection of the matching window depends on the region type, the adaptive identification of the region type is the prerequisite for maximizing the matching entropy term.

\subsubsection{Indicator for occluding and occluded regions}
 Since one pixel of the central view image corresponds to one scene point, we classified the pixels of the center view image into four types. In consequence, the central view image can be divided into occluding, occluded, textured, and smooth regions, which are labeled as $\Omega_\text{occluding}$, $\Omega_\text{occluded}$, $\Omega_\text{texture}$ and $\Omega_\text{smooth}$, respectively. The region indicator function is denoted as $I(x,y)$.
\begin{equation}
		I(x, y)=\left\{\begin{array}{ll}
			0, & (x, y) \in \Omega_{\text {occluding}} \\
			1, & (x, y) \in \Omega_{\text {occluded}} \\
			2, & (x, y) \in \Omega_{\text {texture}} \\
			3, & (x, y) \in \Omega_{\text {smooth}}
		\end{array}\right.
		\label{eq8}.
	\end{equation}

\begin{figure}[ht!]
		\centering\includegraphics[width=12cm]{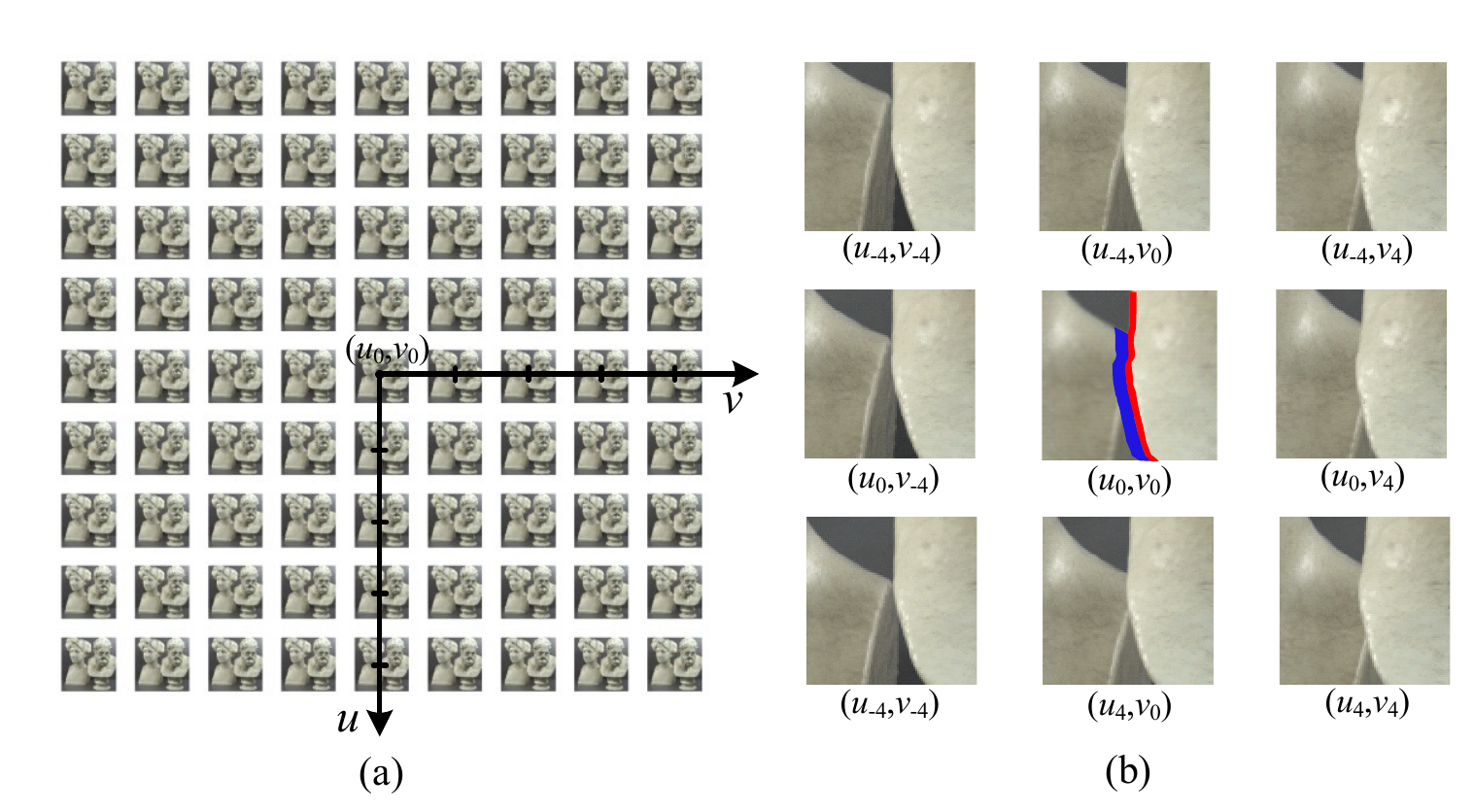}
\caption{ (a) $ 9 \times 9$ sub-aperture images array of the Greek scene's light field. (b) The close-ups of sub-aperture images visually display occluding and occluded pixels.}
		\label{fig3}
\end{figure}

A sub-aperture image array is a visualization of 4D light field data, as shown in Fig.~\ref{fig3}. By varying the viewpoints, the occluding and occluded regions in the images will change. The occluding region is the edge area of the object that causes occlusion. Therefore, the light it emits is visible from all viewpoints. The occluded region is the edge area of the occluded object, and the light it emits is not visible from some viewpoints.

Since occlusion only exists around the edges of objects, occluding and occluded regions can be identified by the differences between the segmentations of the central view image and other sub-aperture images. Considering the farthest sub-aperture images from the central view in eight directions, the occluding and occluded regions can be determined according to the differences. Let $\tilde{\phi}$ be the viewpoint index set of the nine sub-aperture images used to indicate occlusion.

Applying alpha matting\cite{4359322} to the sub-aperture images ${L}_{{({u}}_{i},{v}_{j})}\left({x},{y}\right)$ and ${({u}}_{i},{v}_{j})\in{\tilde{\phi}}$, we can obtain the segmentations ${M}_{\left({u}_{i},{v}_{j}\right)}\left({x},{y}\right)$, as shown in Fig.~\ref{fig5}(b).  If a scene point is visible in the central view, but not visible in some views, then the scene point is occluded, and its corresponding pixel is an occluded pixel. On the contrary, the scene point in front that occludes the occluded scene point, the corresponding pixel is an occluding pixel. In other words, the occluding pixels are occluding in front, while the occluded pixels are occluded in back. For the segmented image $M_{(u_0,v_0)}(x,y) $ of the central view and the segmented image $M_{(u_i,v_j)}(x,y) $ of the $(u_i,v_j)$ view, the closer the object is, the greater the pixel value in the segmentation image. Occlusion occurs at the edge of the object, therefore, for occluding pixels $(x,y) \in \Omega_{\text {occluding}}$, $M_{(u_0,v_0)}(x,y) > M_{(u_i,v_j)}(x,y) $. For the occluded pixel $(x,y) \in \Omega_{\text {occluded}}$, $M_{(u_0,v_0)}(x,y) < M_{(u_i,v_j)}(x,y) $. For non-occlusion cases, $M_{(u_0,v_0)}(x,y) = M_{(u_i,v_j)}(x,y)$. 

The difference between the central view segmentation ${M}_{\left({u}_\mathbf{0},{v}_\mathbf{0}\right)}\left({x},{y}\right)$ and other segmentations is calculated as
\begin{equation}
		diff(x, y)=\sum_{(i, j) \in \tilde{\phi}}\left(M_{\left(u_{0}, v_{0}\right)}(x, y)-M_{\left(u_{i}, v_{j}\right)}(x+\Delta u, y+\Delta v)\right)
\label{eq9},
\end{equation}
where $\Delta{u}=u_i-u_0$, $\Delta{v}=v_j-v_0$. 

As a result, the occluded regions ${\Omega}_{\text{occluded}}$ 
and the occluding regions ${\Omega}_{\text{occluding}}$ can be identified.
\begin{equation}
		\Omega_{\text {occluded }}=\{(x, y) \mid  diff(x, y)<0\}
		\label{eq10},
\end{equation}
\begin{equation}
		\Omega_{\text {occluding }}=\{(x, y) \mid  diff(x, y)>0,  (x, y) \notin \Omega_{\text {occluded}}\}
		\label{eq11}.
\end{equation}

 From Eq.\ref{eq11}, the occluding pixels are defined to be visible under all viewpoints. Therefore, as long as the scene point is occluded, the corresponding pixel is the occluded pixel. The object in the middle depth layer will block the object points behind it at some viewpoints and may be blocked by object points in front of it at some viewpoints.

\subsubsection{Indicator for textured and smooth regions }
In smooth regions, the intensities of pixels in a local neighborhood tend to be similar. Therefore, the statistical intensity characteristics in the neighborhood can be used to measure the smoothness, and then the smooth region ${\Omega}_{\text{smooth}}$ can be identified as
\begin{equation}
		\Omega_{\text {smooth }}=\left\{(x, y) \mid \psi(x, y)<\tau,(x, y) \notin \Omega_{\text {occluded }} \cup \Omega_{\text {occluding }}\right\}
		\label{eq12}.
	\end{equation}
where ${\psi}\left({x},{y}\right)$ is the number of pixels within the neighborhood that have different pixel values from $\left({x},{y}\right)$. In this paper, we choose the empirical parameter $\tau = \frac{N}{2}$, where $N$ is the number of pixels within the neighborhood.

\begin{figure}[ht!]
		\centering\includegraphics[width=13cm]{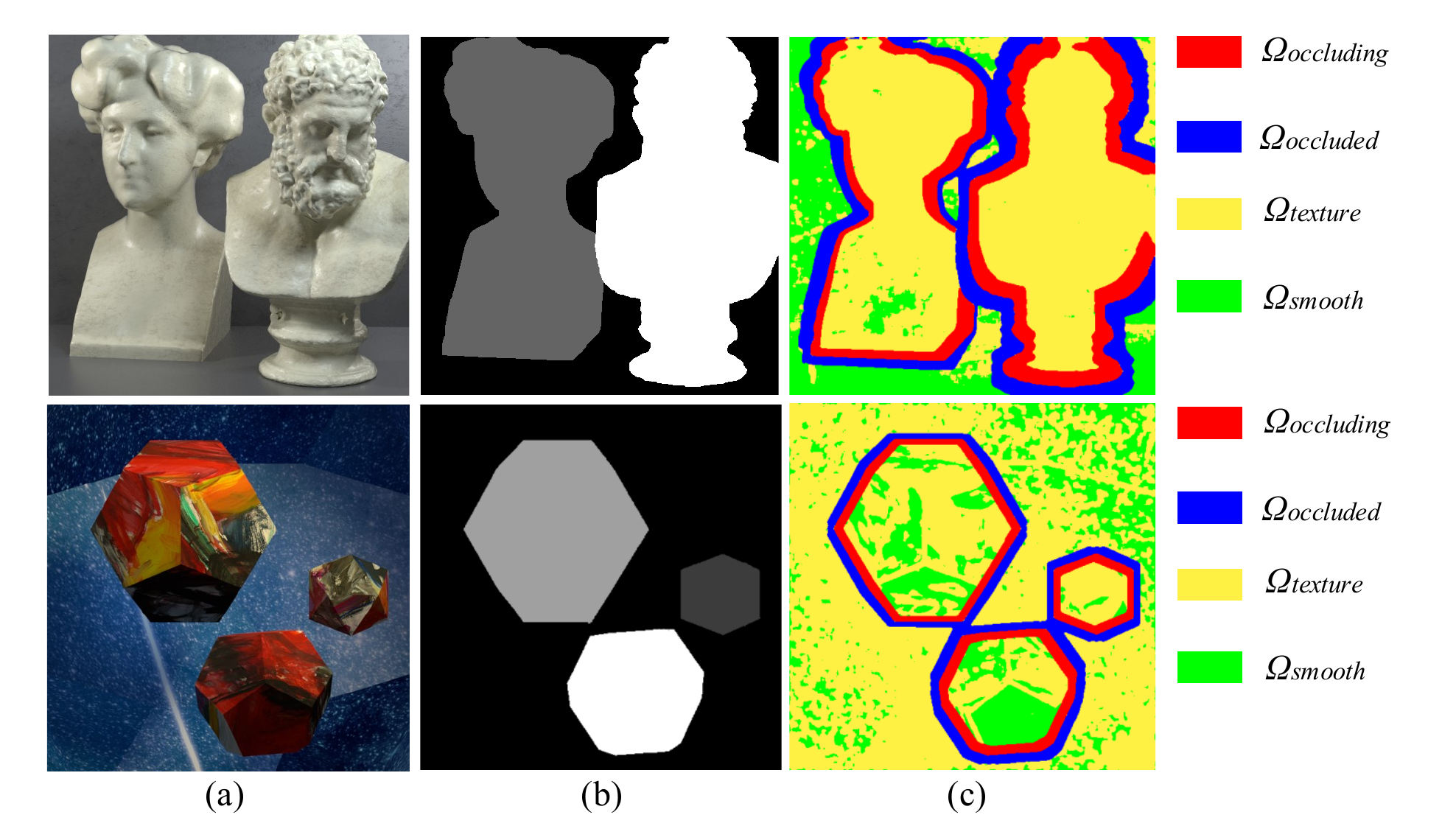}
\caption{Results of adaptive identifying the region types for Greek scene and Platonic scene. (a) The central view image. (b) The segmentation of the center view. (c) The region types. The occluding, occluded, textured, and smooth regions are marked as the red pixels, the blue pixels, the yellow pixels, and the green pixels respectively in (c).}
		\label{fig5}
\end{figure}

After the occluding, occluded, and smooth regions are identified, the remaining regions are the textured regions ${\Omega}_{\text{texture}}$.  Both occluding regions and occluded regions can be further classified into textured or smooth regions. In our disparity estimation process, both shape and size of the matching windows are carefully selected for occlusion regions. For textured and smooth regions, only the size of the matching windows are considered. Therefore occlusion regions do not need to be further classified. In our postprocessing refinement, only smooth regions are refined with TV regularization to deal with 'black holes'. Since 'black holes' rarely appear around edges where occlusion may occur, the further classification of the occlusion region is also not necessary in postprocessing. As a result, we classify the regions into four types of regions with no overlap. Taking the Greek scene and the Platonic scene as examples, the region identification results are shown in Fig.~\ref{fig5}.

\subsection{Adaptive selection of matching window by region type}

Based on the identified region, optimal matching windows can be designed for different regions. The shape and size of the windows are obtained by maximizing the matching entropy term. In this work, the disparity consistency part of the matching entropy term is calculated by the initial disparity map extracted from the images in the viewpoint set $\tilde{\phi}$.

\subsubsection{Matching window selection and visible viewpoint set adoption for occlusion regions}
For anti-occlusion in area matching, the matching window of occluding pixels should not contain occluded pixels, while the matching window of occluded pixels should not contain occluding pixels. As a result, the key for selecting a matching window with high matching entropy is to find the effective shape of the matching window and the visible viewpoint set to exclude pixels in the opposite occlusion situation. 

\begin{figure}[ht!]
\centering\includegraphics[width=10cm]{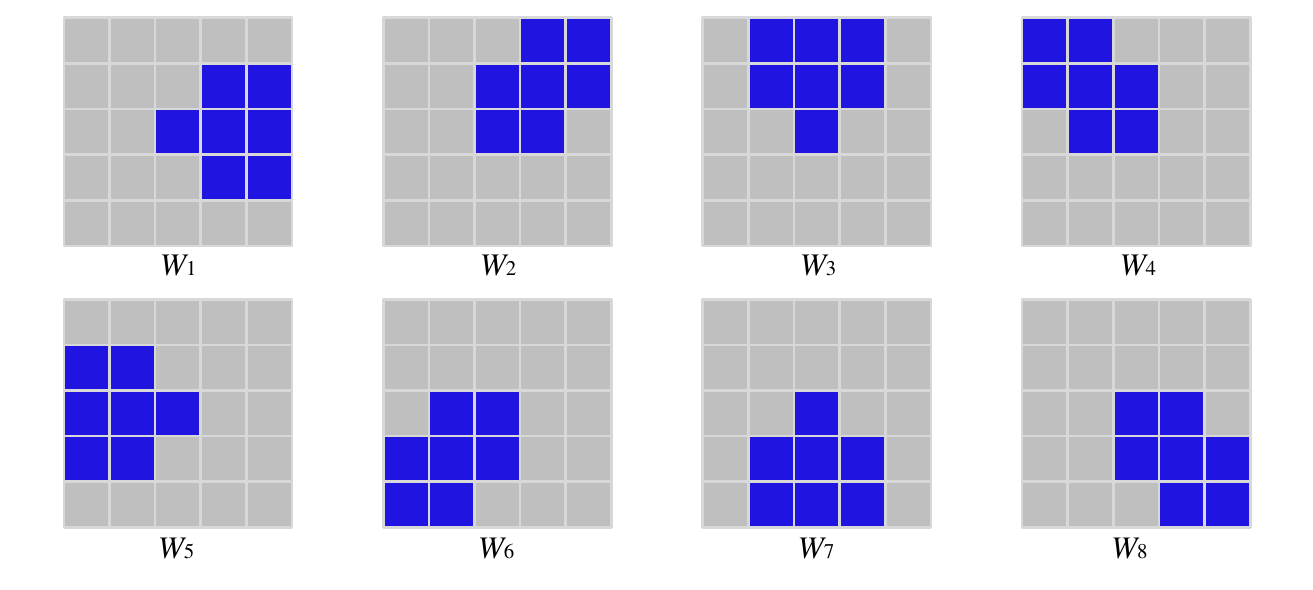}
\caption{ The preset matching window shapes. The 8 preset shapes are marked as blue pixels.}
		\label{fig6}
\end{figure}

To select matching windows for occlusion regions, we should determine the shape and the size of each window. By considering the directions of the occlusion, we preset the eight window shapes $W_i (i=1,\ldots,8)$ as shown in Fig.~\ref{fig6}. We also preset the size range of the window from ${3}\times{3}$ to ${15}\times{15}$. Then the optimal shape and size can be searched from the preset shapes and sizes to reach the maximum matching entropy. 

\begin{figure}[ht!]
\centering\includegraphics[width=13cm]{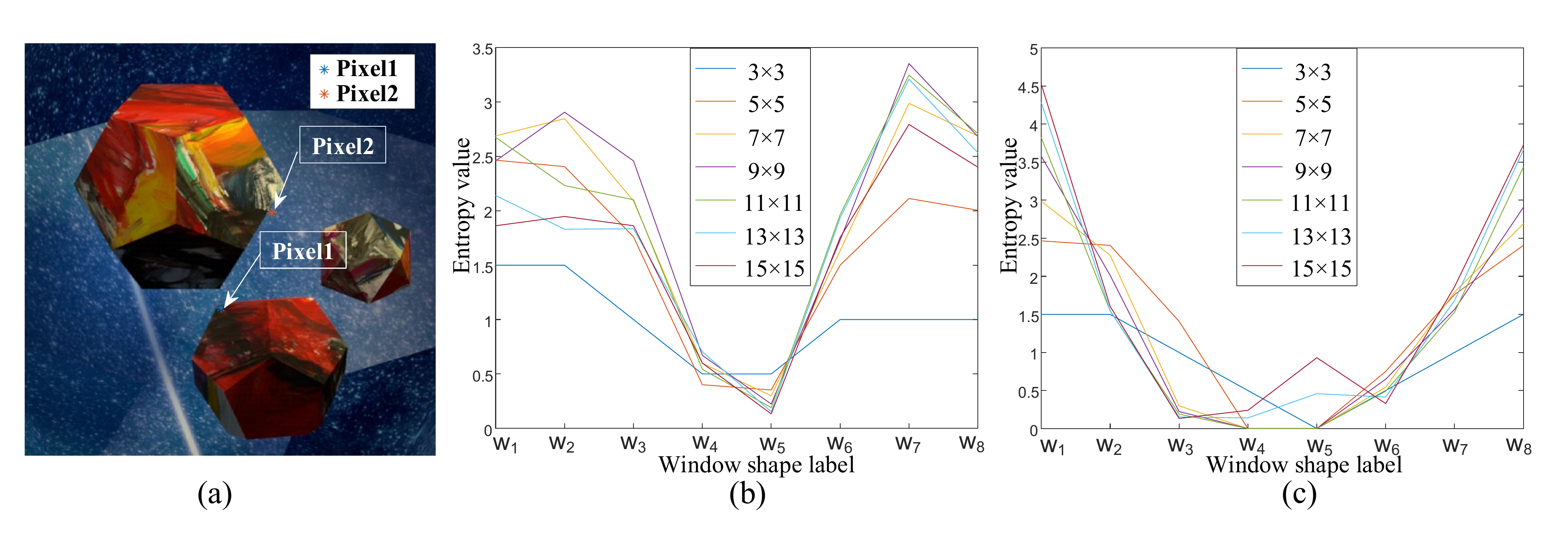}
\caption{(a) Pixel1 and Pixel2 are the representative pixels for the occluding pixel and the occluded pixel, respectively. (b) and (c) show the change curves of matching entropy as the window changes for Pixel1 and Pixel2, respectively. The colors of the curves represent windows of different sizes.}
\label{fig7}
\end{figure}

Taking the occluded Pixel1 in Fig.~\ref{fig7}(a) as an example, Fig.~\ref{fig7}(b) shows the relationship between the matching entropy values and the preset matching windows. The $X$-axis is the window shape labels, and the $Y$-axis is the matching entropy value. Different line colors represent different window sizes. For Pixel1, the maximum matching entropy value is on the purple line, whose $X$-axis is at 7, which means the optimal window should be a $ W_{7}$ window with a size of ${9}\times{9}$.


\begin{figure}[ht!]
		\centering\includegraphics[width=10cm]{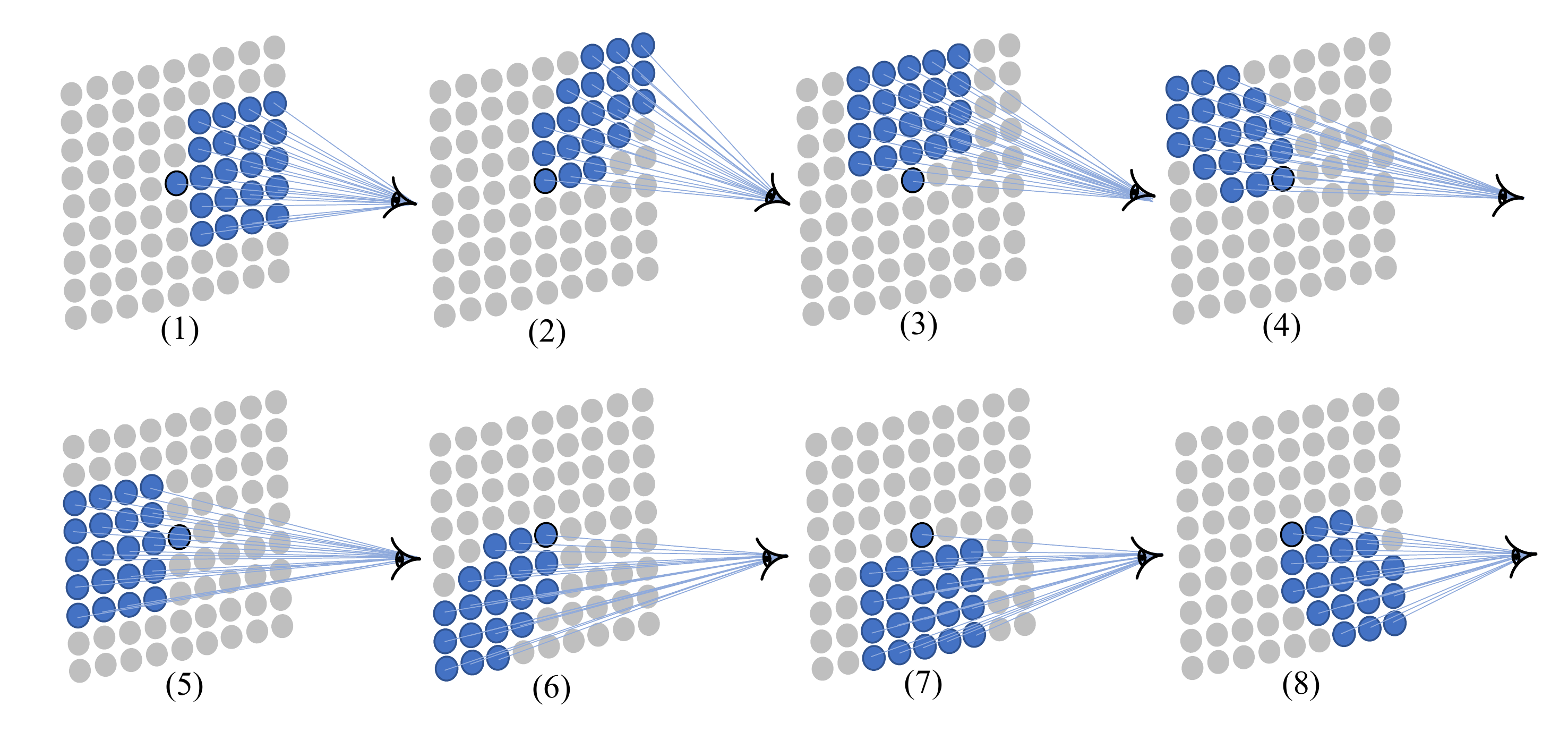}
\caption{ The set of preset visible viewpoints. The blue viewpoints represent the visible viewpoints.}
		\label{fig9}
\end{figure}

For smooth, textured, and occluding regions, the windows are visible from all viewpoints; Thus, $\phi$ in these regions should be a complete viewpoint set. For occluded regions, some pixels in the window are not visible from some viewpoints, so it is necessary to eliminate these viewpoints from the complete viewpoint set. For each window shape, there is a corresponding visible viewpoint set in Fig.~\ref{fig9}. For instance, the pixels in the window $W_1$ are visible under the viewpoints in Fig.~\ref{fig9}(1).

\begin{figure}[ht!]
		\centering\includegraphics[width=13cm]{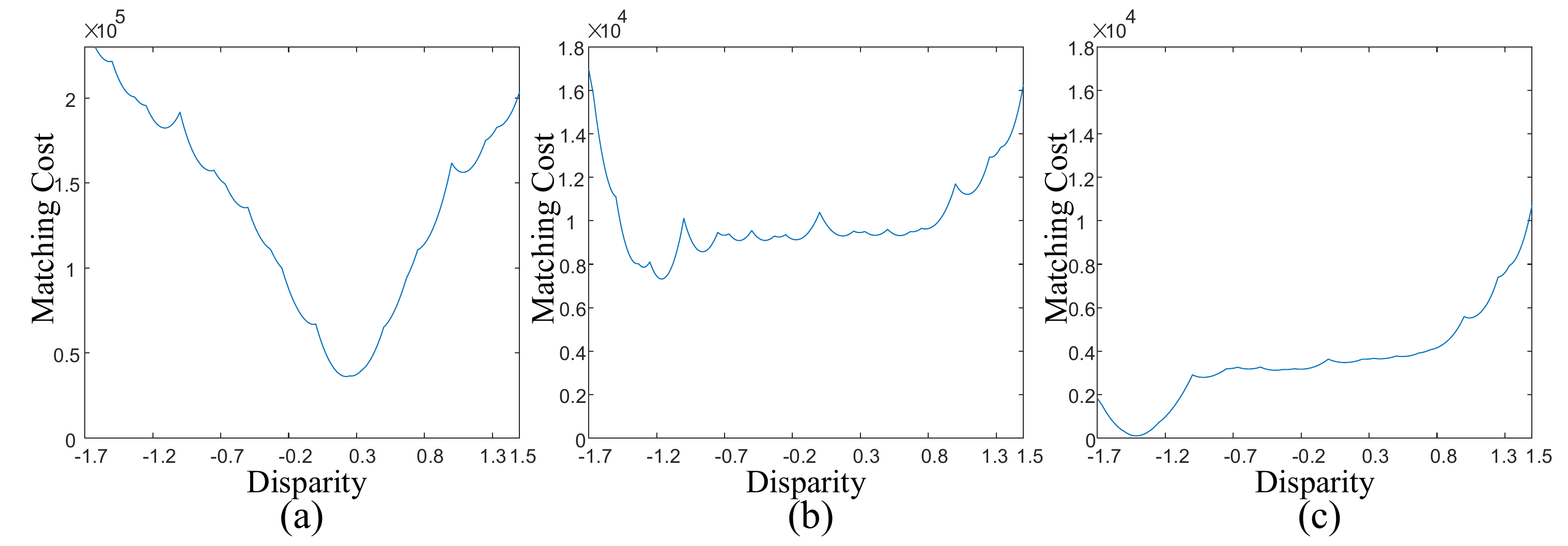}
\caption{The relationships between the matching costs and the disparity values via different matching windows and viewpoint sets. (a) The fixed square matching window with the complete viewpoint set. (b) The adaptively selected window with the complete viewpoint set. (c) The adaptively selected window with the visible viewpoint set.}
		\label{fig10}
\end{figure}

To verify the effectiveness of the selection of the visible viewpoint set for occluded pixels, we calculate the matching cost of Pixel2 in Fig. \ref{fig7} under different disparity values with the traditional fixed matching window, adaptive matching window and adaptive matching window with the visible viewpoint set. The relationships between the disparity values and matching costs are denoted by the curves in Fig. \ref{fig10}. The ground truth is $-1.42$, and the minimum point in (c) leads to the most accurate disparity value.

\subsubsection{Matching window selection for smooth and textured regions }
In smooth and textured regions, the main consideration of the matching window design is to make the window size cover effective texture richness. Taking the matching entropy and computational cost into consideration, the optimal window size is searched from $3\times3$ to $15\times15$.

\begin{figure}[ht!]
		\centering\includegraphics[width=13cm]{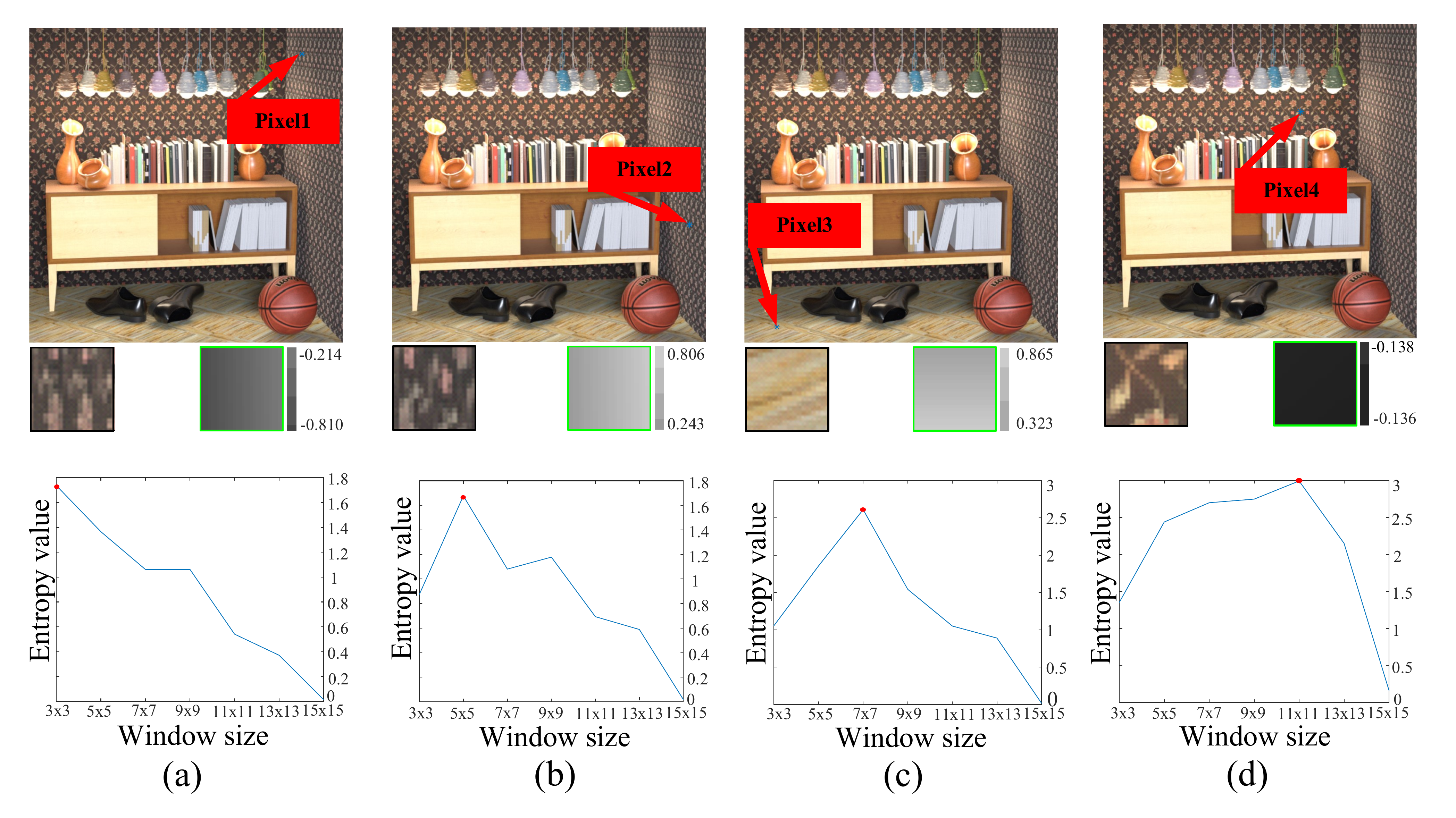}
\caption{ Four typical matched pixels with different disparity consistencies and the relationship between the matching entropy value and the window size.
 The typical matched pixels in the local region and the corresponding disparity map are shown in close-ups mode. The Pixel4 is on a flat background, and there is no difference in the disparity value in its neighborhood, so Pixel4 has the best disparity consistency. Pixel1 and Pixel2 in the scene are on the same wall. Pixel1 is near, while Pixel2 is far away, so Pixel2 has better disparity consistency than Pixel1.}
		\label{fig8}
\end{figure}

In Fig.~\ref{fig8}, we select four pixels with different disparity consistencies from a neighborhood to verify the effectiveness of the designed matching window. The second row of Fig.~\ref{fig8} shows the relationship between the matching entropy value and the matching window size. The $X$-axis is the window size, and the $Y$-axis is the matching entropy value. Pixel1 is the least consistent pixel, and Pixel4 is the most consistent pixel. From the changing curve, the optimal window size for Pixel1 is ${3}\times{3}$, and the optimal window size for Pixel4 is ${11}\times{11}$.
	
\subsection{Disparity estimation and postprocessing }
\subsubsection{Disparity estimation and refinement }
After adaptively selecting the optimal matching window and determining the visible viewpoint sets, a disparity map can be estimated by minimizing the objective functional. Since the effective information of the matching windows in smooth regions may be insufficient, the smooth regions are shown as “black holes” in the disparity map. The total variation(TV) model\cite{1992Nonlinear}  is used in the smooth region to eliminate the 'black holes' and refine the disparity value $disp^\ast\left(x,y\right)$.

\begin{equation}
		\mu(x, y)=\arg \min _{\mu}\left\{\iint_{\Omega_{\text{smooth}}} \left(\mu(x, y)-{disp}^{*}(x, y)\right)^{2} d x d y+\gamma|\nabla \mu(x, y)| d x d y\right\}
		\label{eq13}.
	\end{equation}
where $\mu\left(x,y\right)$ is the disparity map refined by the TV model, ${\nabla\mu}\left(x,y\right)$ is the gradient of the disparity $\mu\left(x,y\right)$, and $\gamma$ is the regularization parameter.

In addition, the line search method provides jagged estimation results. The TV model can reduce the jaggedness but blurs the edges at the same time. Therefore, we only apply TV in smooth regions. 
\section{Experimental results}
In this section, experiments on both synthetic and real data are performed to evaluate the effectiveness of the proposed method. We compare the result of our method with those of five other methods on the 4D Light Field Benchmark Dataset\cite{8014960}. The real data are acquired by a camera with a three-axis translation platform.  Our method was implemented in MatLab2020b. We set $\alpha_1=1$,  $\lambda=1$, $\gamma=0.2$.  $\alpha_2=1$ 
 in occluding and occluded regions, while $\alpha_2=0$ 
 in textured and smooth regions.

\subsection{Evaluation and comparison of the algorithms}
In the 4D Light Field Benchmark dataset, the light field data provided for each scene are a $9\times9$ sub-aperture image arrays (with spatial resolutions of $512\times512$). We compared the performance of our method with that of five state-of-the-art methods: LF\cite{2015Accurate}, epi1\cite{2016What}, LF\_OCC\cite{2016Occlusion}, MV\cite{8014960}, and mvcmv0\cite{8014960}.

\begin{figure}[ht!]	\centering\includegraphics[width=13cm]{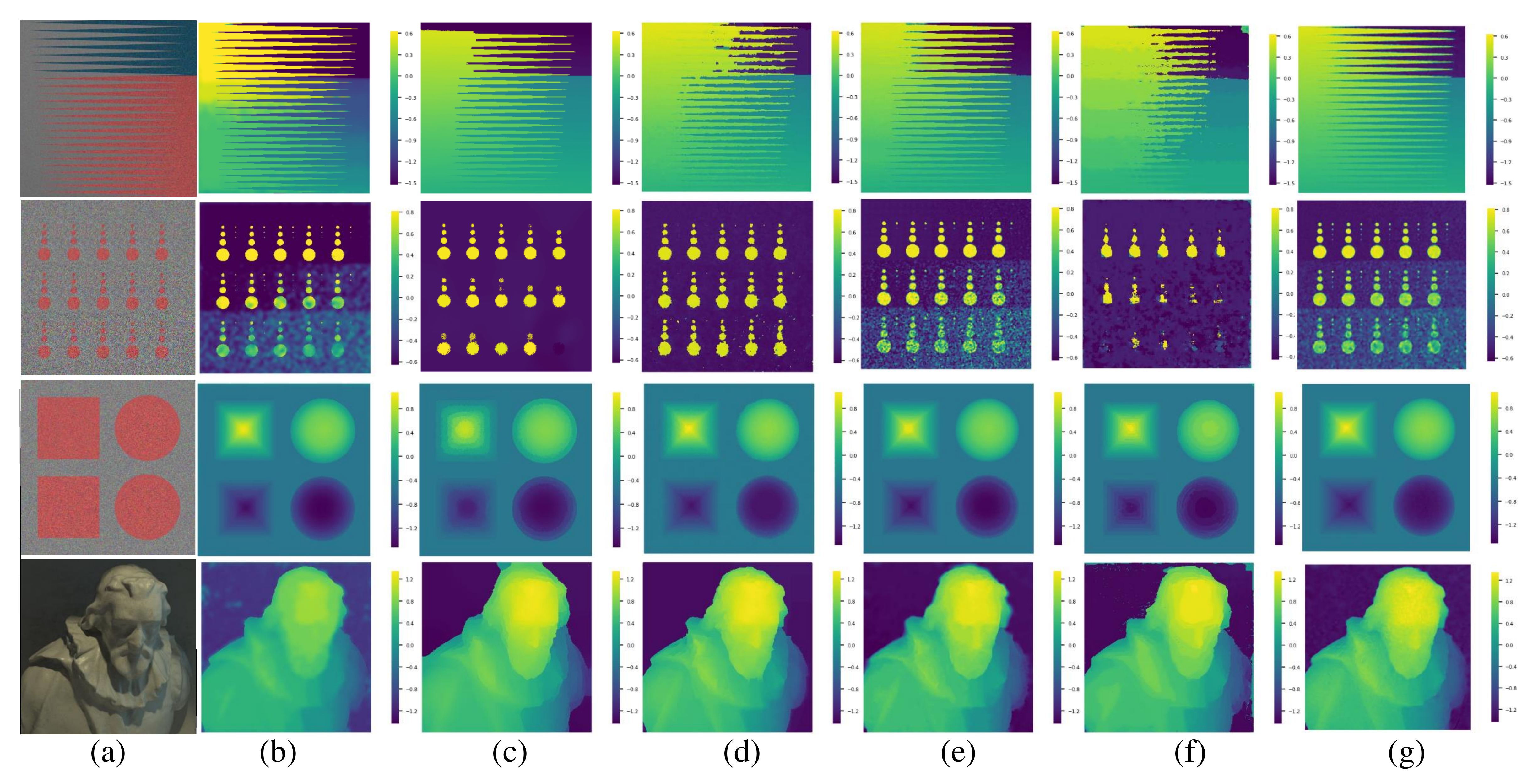}
\caption{ Backgammon scene (1st row), Dots scene (2nd row), Pyramid scene (3rd row), and Cotton scene (4th row). (a) Central view image, (b) Our result, (c) LF, (d) LF\_OCC, (e) MV, (f) mvcmv0, (g) epi1. The results of (c)-(g) come from\cite{8014960}.}
\label{fig11}
\end{figure}

Four different scenarios are selected from the dataset to evaluate the performance of the proposed method, as shown in Fig.~\ref{fig11}. The Backgammon scene is designed to assess the interplay of fine structures, occlusion boundaries, and disparity differences. The Dots scene is designed to assess the effect of camera noise on the estimation of objects of varying sizes. The Pyramid scene is used to evaluate the performance of the algorithm in convex, concave, circular, and planar geometries. The Cotton scene is closer to a real scene with less artificial design and is used to evaluate the estimation accuracy in smooth and textured regions.

In Fig.~\ref{fig11}, the disparity estimation results of the Backgammon scene show that LF\_OCC and mvcmv0 produce noise at edges, epi1 loses edge details, and LF produces a certain degree of blur at the gap between jagged areas. In contrast, our method and MV can maintain fine edge information and gap structures. The results of the Dots scene show that our method is robust to noise. Our method, MV and epi1, can estimate more dot structures when the noise level is high. The results of the Pyramid scene show that our method and epi1 obtain smoother results on convex and concave inclined planes. For the Cotton scene, the estimation result of our method is blurred to a certain extent, but more details are maintained at the boundary of the foreground.

We thoroughly assess and compare the six methods by summarizing all scores computed for each scene and the associated metrics into a radar chart. Each radar axis represents one metric, where the zero in the center represents perfect performance. Backgammon Thinning and Backgammon Fattening are used to evaluate the preservation of the fine structures in the Backgammon scenes. The Dots Missed Dots and the Dots Background mean square error (MSE) metrics are used to evaluate the anti-noise performance in the Dot scene. The Pyramids Bump Slanted and Pyramids Bump Parallel metrics are used to evaluate the smoothness of the slanted and parallel planes in the Pyramid scene. The MSE is the median MSE of the four scenarios and indicates the comprehensive performance of each method.
\begin{figure}[ht!]	\centering\includegraphics[width=10cm]{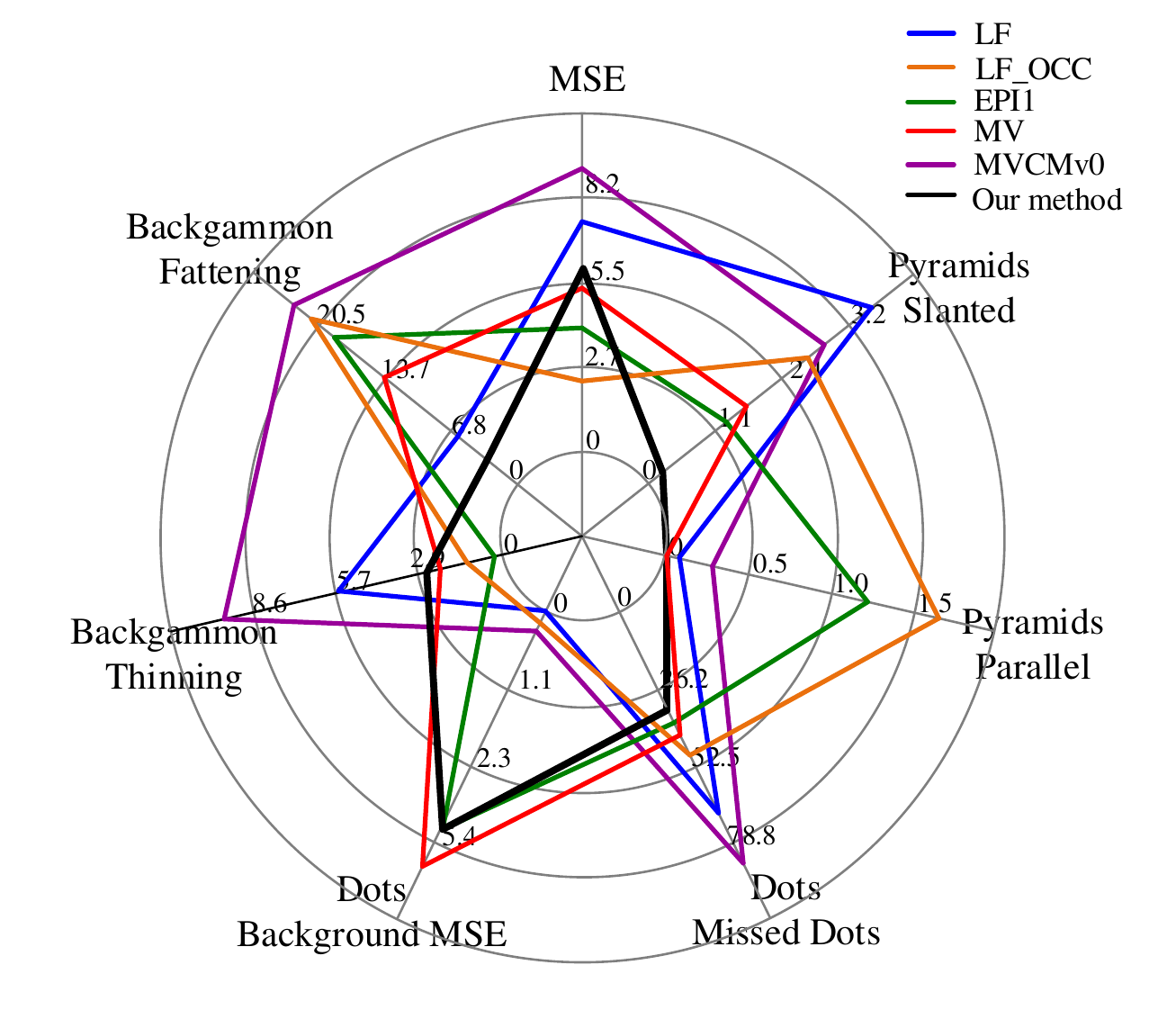}
\caption{ The radar chart on the evaluation scenes. Our proposed matching entropy based method has advantages on several indicators.}
		\label{fig12}
\end{figure}
The comprehensive performance of our method is at the middle level according to the MSE score. In terms of Pyramids Bump Slanted, Pyramids Bump Parallel, Dots Missed Dots, and Backgammon Fattening, our method performs best. In terms of Backgammon Thinning, our method is mediocre. In terms of Dots Background MSE, the performance is weak. The overall performance indicates that our method can estimate more accurate disparity maps in smooth and textured regions and can handle occlusion regions well when the noise levels are high. Furthermore, with increases in the noise and fineness levels, the estimation accuracy of the background is more likely to decrease than that of the foreground structure.

\subsection{Error analysis in occlusion regions}
\begin{figure}[ht!]
\centering\includegraphics[width=13cm]{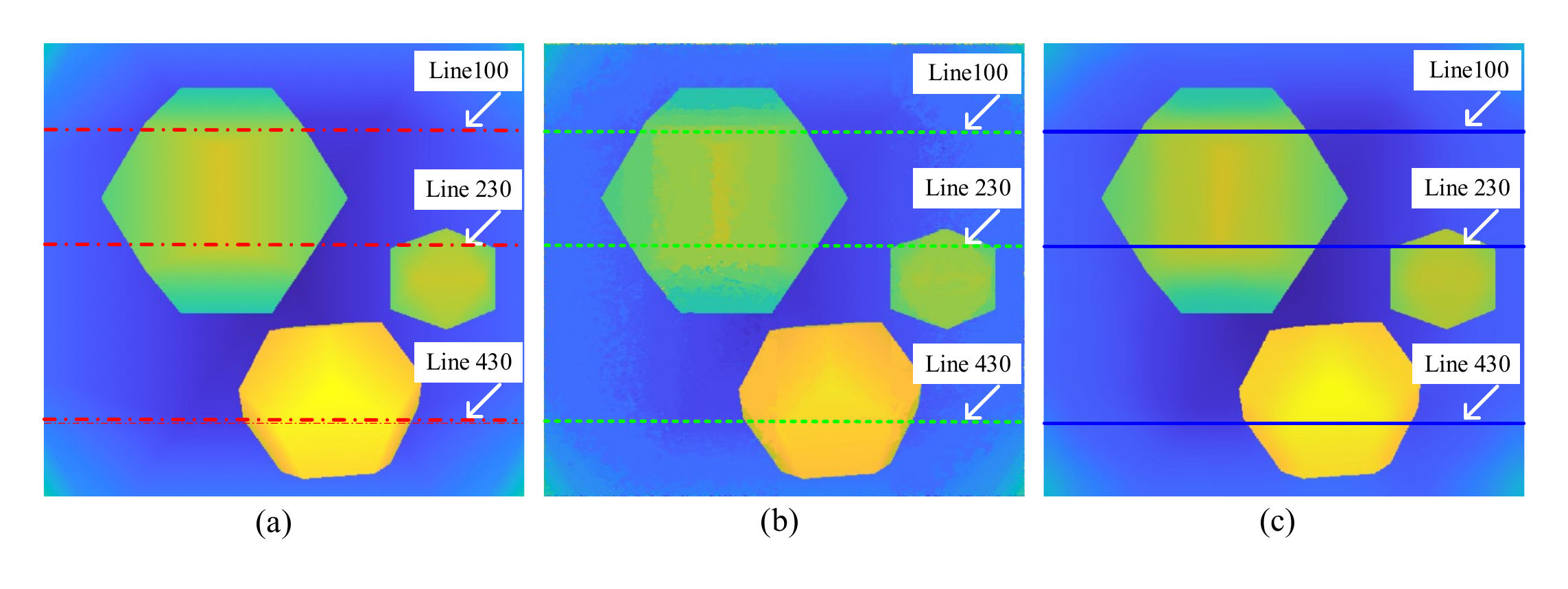}
\caption{The results of Platonic scene. (a) Ground truth of the disparity. (b) Initial disparity map. (c) Refined disparity map. And the positions of the profiles are marked, i.e, Line 100, Line 230, and Line 430.}
\label{fig13}
\end{figure}

Occlusion exists around the edges of the scene, and the blurriness of the edges can reflect the estimation accuracy of the occlusion regions. To further evaluate the effectiveness of our method in occlusion and smooth regions, we analyze the Platonic scene in detail by drawing profiles to compare the ground truth disparity map and the calculated disparity map, as shown in Fig.~\ref{fig13}. The profile positions are selected to include as many occlusion regions as possible, and we focus on the accuracy of the estimation results in the occlusion regions.
\begin{figure}[ht!]
\centering\includegraphics[width=13cm]{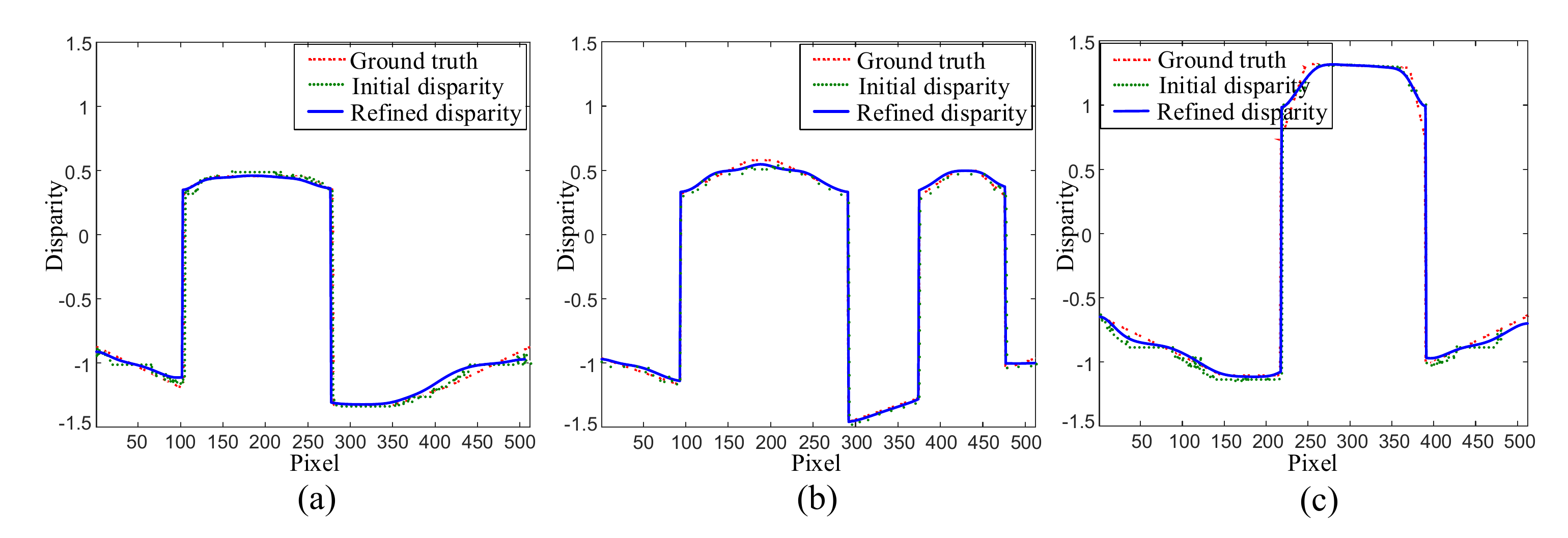}
\caption{Profiles of the disparity maps. (a) Profiles of Line 100. (b) Profiles of Line 230. (c) Profiles of Line 430. The red line represents the ground truth disparity, the green line represents the initial disparity, and the blue line represents the refined disparity.}
\label{fig14}
\end{figure}

Gaps in the disparity values will occur crossing object edges.  In Fig.~\ref{fig14} (a) - (c), it can be seen that the initial disparity and the refined disparity values jump at the position where the ground truth disparity values jump. This shows that our method can maintain the edges well and can accurately estimate disparity values in occlusion areas. In addition, the profile is closer to the ground truth with fewer jagged areas and false jumps after TV refinement.

\begin{figure}[ht!]
\centering\includegraphics[width=13cm]{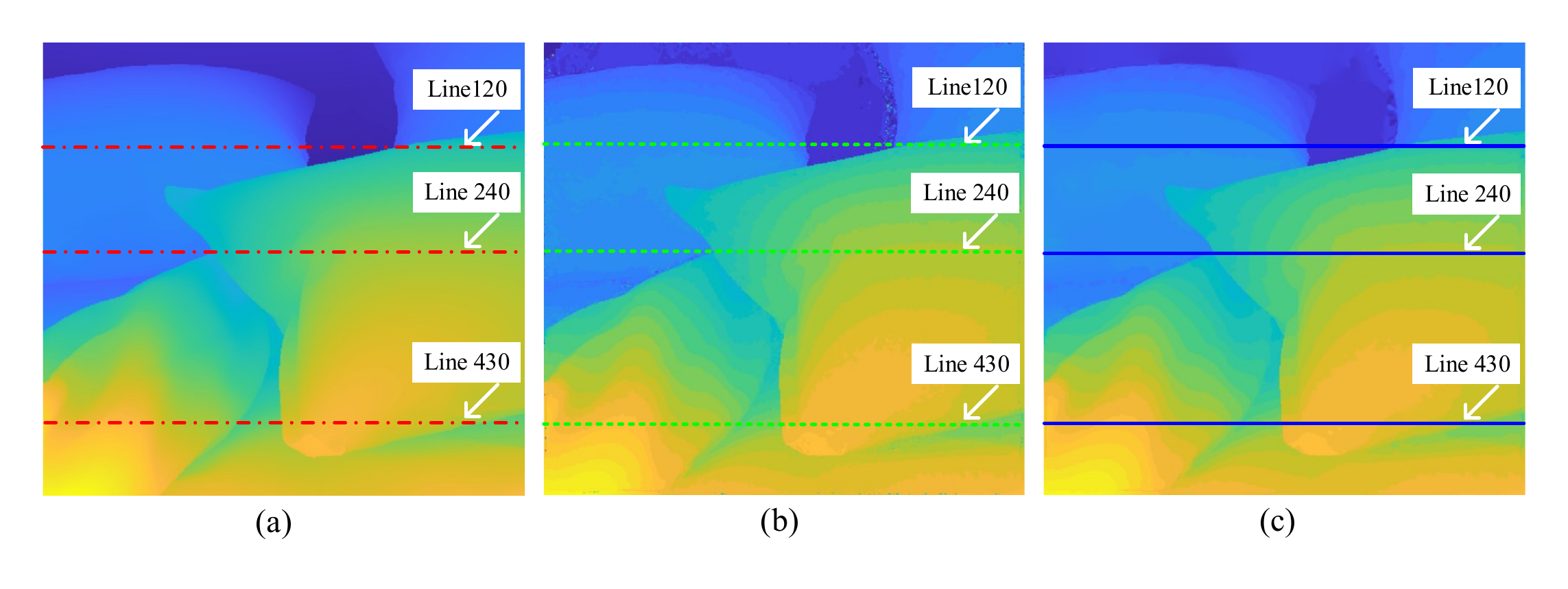}
\caption{The results of the Pillows scene. (a) Ground truth of the disparity. (b) Initial disparity map. (c) Refined disparity map. And the positions of the profiles are marked in Line 120, Line 240, and Line 430.}
\label{fig23}
\end{figure}

\begin{figure}[ht!]
\centering\includegraphics[width=13cm]{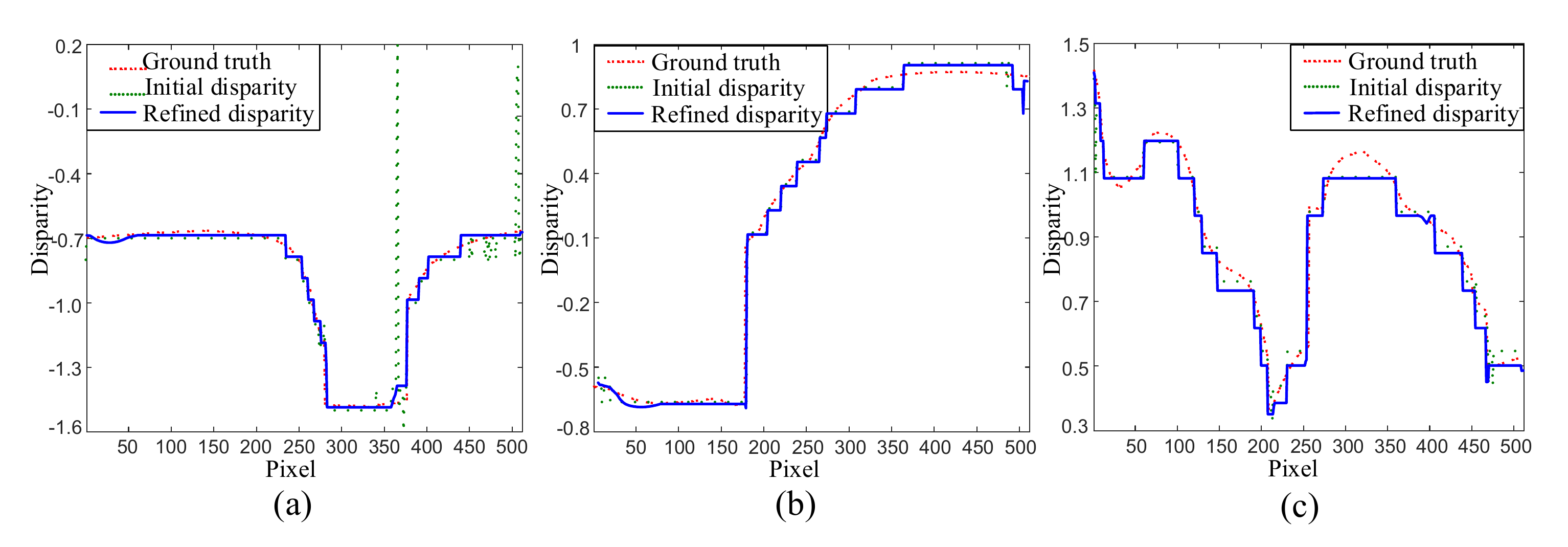}
\caption{Profiles of the disparity maps. (a) Profiles of Line 120. (b) Profiles of Line 240. (c) Profiles of Line 430. The red line represents the ground truth disparity, the green line represents the initial disparity, and the blue line represents the refined disparity.}
\label{fig24}
\end{figure}

In order to further analyze the effectiveness in complex occlusion scenarios, we adopt the pillows scene as shown in Fig.~\ref{fig23}. We use the profiles of the disparity map (as shown in Fig.~\ref{fig24}) to analyze the occlusion processing results of the proposed method. In Fig.~\ref{fig23}, both Line 120 and Line 240 pass through the complex occlusion regions. In Fig.~\ref{fig23}, it can be seen from (a) and (b) that the jump positions of the blue line and the red line are the same, indicating that the proposed method is effective in the complex occlusion situations.


\subsection{Experiments on real data}
In the experiment, a camera controlled by a three-axis translation platform is used to collect 2D images from $9\times 9$ different viewpoints uniformly spaced in a plane at an interval of 0.5 mm to obtain 4D light field data. The resolution of the detector is $1280\times980$, and the focal length of the lens is 35 mm.

The effectiveness of our method at the disparity gaps is verified in the first real data experiment, where four standard cubes are placed at four different disparities in a range of $[90, 100]$ cm. The effectiveness of our method when the disparity changes continuously is verified in the second real data experiment, where regular-shaped blocks (standard pyramid, cuboid, cone, hemisphere, and cylinder) are placed in the disparity range of $[90, 100]$ cm. The effectiveness of our method and the benefit of refinement in smooth and occlusion regions in real scenarios is verified in the third real data experiment, where a plant of Tiger Piran is placed at a disparity range of $[85, 100]$ cm.
\begin{figure}[ht!]		\centering\includegraphics[width=13cm]{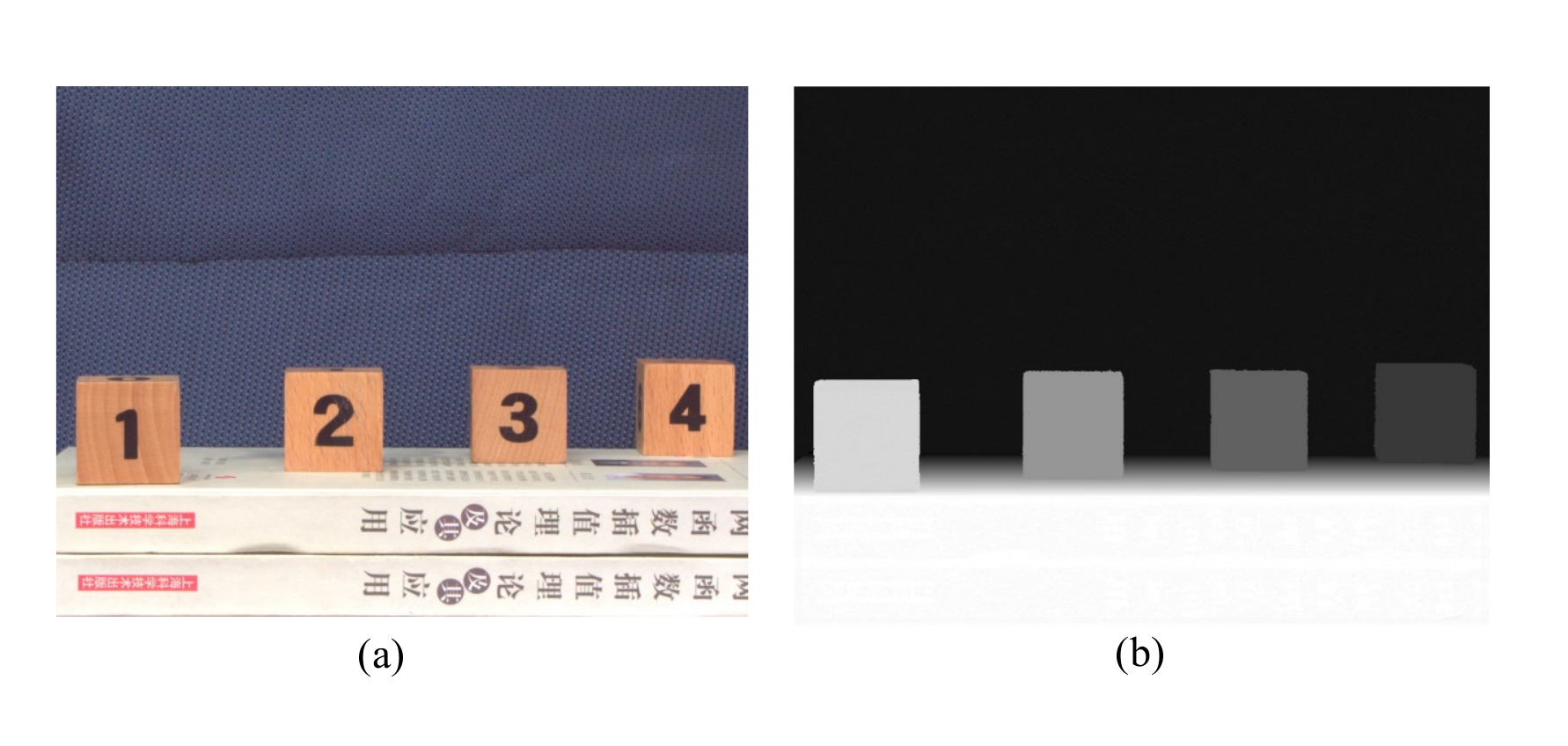}
\caption{The results of the first real data experiment. (a) Central view image of Cubes scene. (b) Disparity map.}
		\label{fig16}
\end{figure}

\begin{figure}[ht!]		\centering\includegraphics[width=13cm]{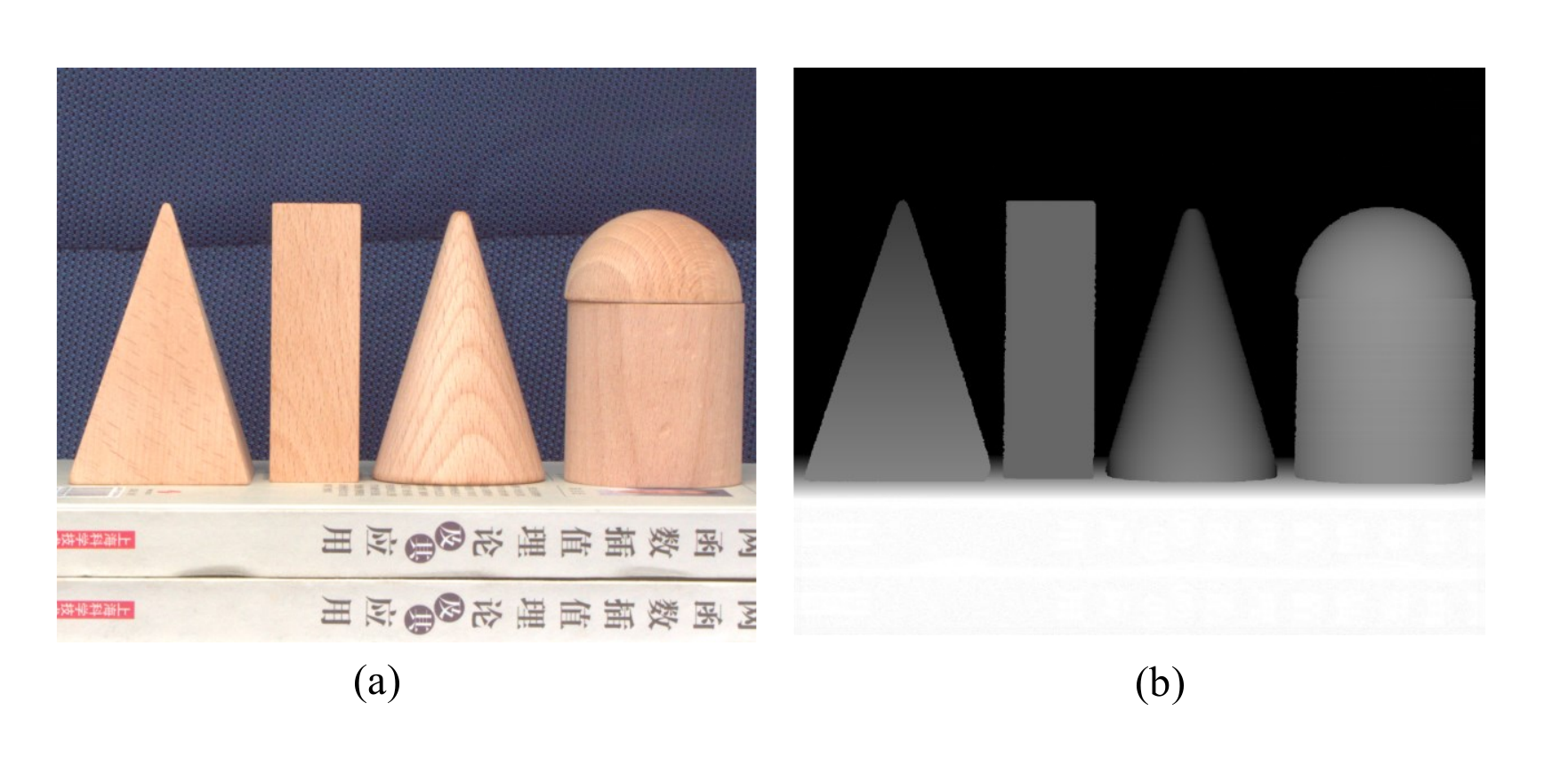}
\caption{The results of the second real data experiment. (a) Central view image of Regular-shaped Blocks scene. (b) Disparity map.}
		\label{fig15}
\end{figure}

Fig. \ref{fig16}(b) shows that the edges of every cube in the estimated disparity map are clear and sharp, which indicates that our method is able to handle disparity gaps well. Fig.~\ref{fig15}(b) shows that when the disparity of the objects changes continuously, our method can maintain continuity in the estimation result.

\begin{figure}[ht!]		\centering\includegraphics[width=13cm]{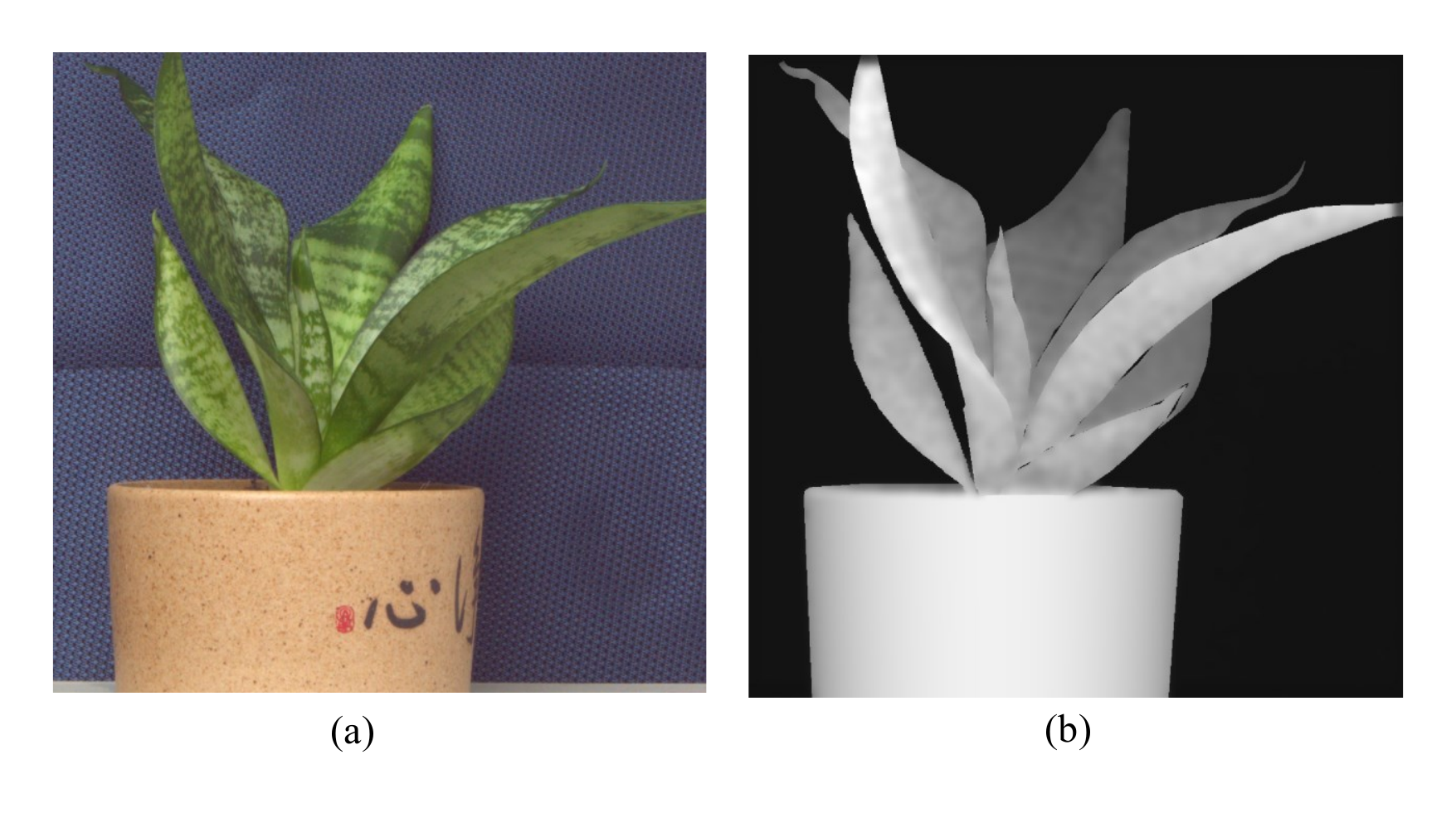}
\caption{The results of the third real data experiment. (a) Central view image of Tiger Piran scene. (b) Disparity map.}
		\label{fig17}
\end{figure}

\begin{figure}[ht!]		\centering\includegraphics[width=13cm]{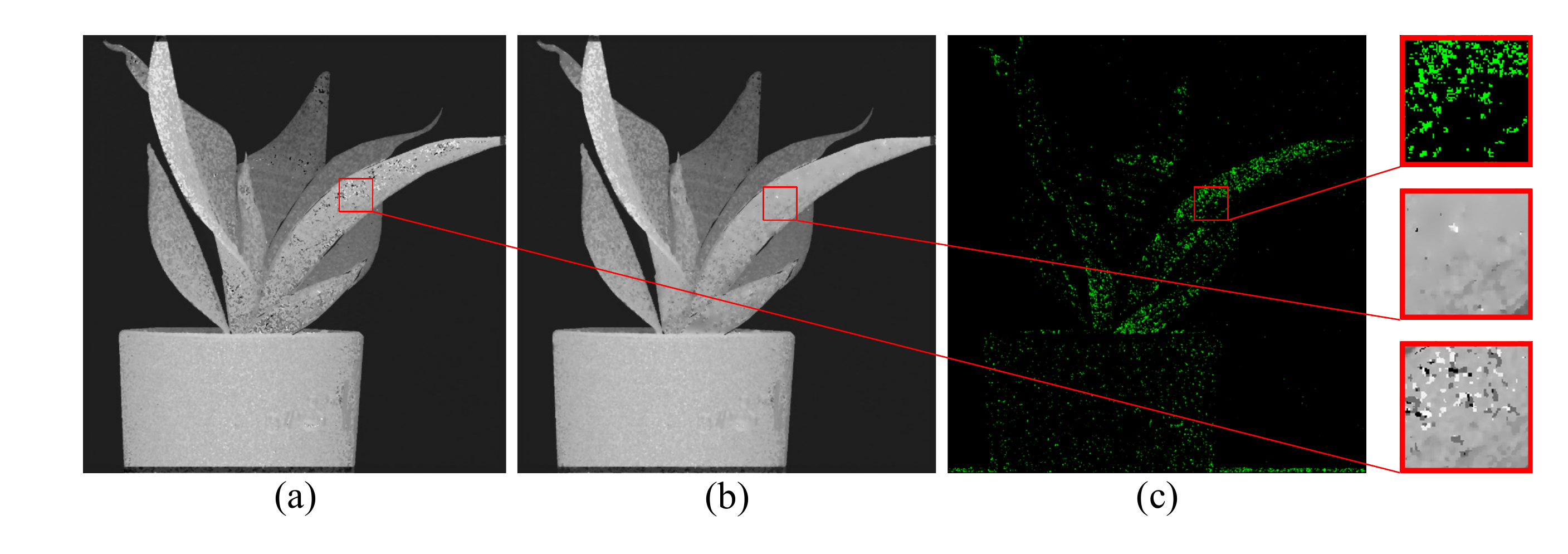}
\caption{Evaluations in the smooth regions. (a) Initial disparity map. (b) Refined disparity map by TV model. (c) Smooth region labeled map. The closed-ups are shown in the right column}
		\label{fig18}
\end{figure}

\begin{figure}[ht!]		\centering\includegraphics[width=13cm]{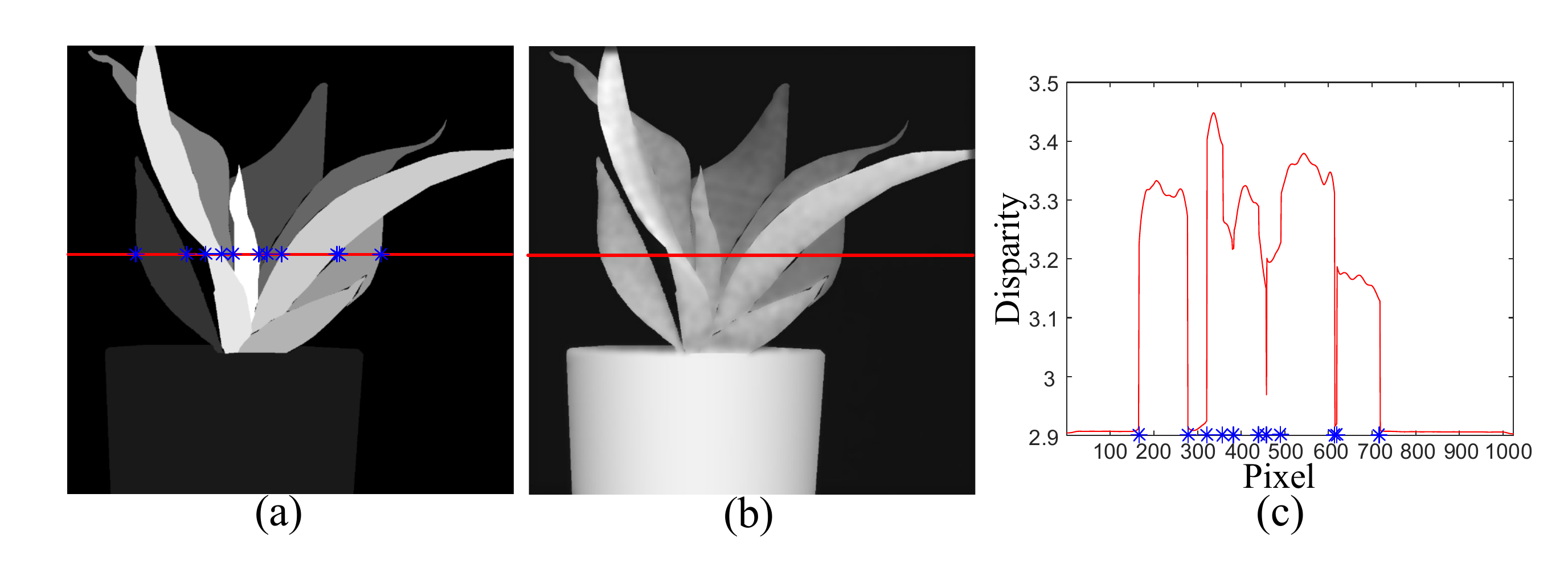}
\caption{Evaluations in the occlusion regions. (a) The segmentation image of the central view image. (b) The estimated disparity map. (c) Profile and edge positions. The position of the profile is marked as the red line in (a) and (b), and the positions of the edges where occlusion occurs are marked as the blue stars.}
\label{fig19}
\end{figure}

Fig. \ref{fig17} shows the result of the third real data experiment.  In the real data experiments, the Tiger Piran scene contains complex occlusion relationships between plant leaves. We can see that our method can produce a disparity map with quite high accuracy for real scenes with smooth, textured, and occlusion regions. To further evaluate the benefit of TV refinement, we focus on the smooth and occlusion regions shown in Fig.~\ref{fig18} and Fig.~\ref{fig19}. In Fig.~\ref{fig18}, we notice that there are "black holes" in the initial disparity map due to the smoothness. The refinement can repair these "black holes" and improve the disparity estimation quality. In Fig.~\ref{fig19}, we use the segmentation map to mark the edges where occlusion occurs. Then, we draw a profile (Line 480) and mark the edge positions on the profile as blue stars. Fig.~\ref{fig19}(c) shows that the jumping positions of the profile coincide with the edge positions. It indicates that the refinement is able to preserve the edge information.  The experiments show that the proposed method can reconstruct the disparity of the occlusion regions in complex situations with multiple layers of occlusion. 

\section{Discussion}
Region matching can be used to effectively estimate disparity information only if matching windows contain sufficiently rich texture information, meet the consistency condition of the initial disparity, and retain only correct matching information when occlusion occurs. To measure the effectiveness of each matching window, the concept of matching entropy is proposed to form the constraint for matching window selection and visible viewpoints set adoption. Considering the segmentation and the local consistency, the region type identification function is constructed. Then, the optimal matching windows and the visible viewpoints in different regions are selected according to the matching entropy value. Finally, the objective functional can be minimized by the line search method, and the high-precision disparity information can be estimated.

To verify the effectiveness of our methods, we conduct experiments with both synthetic and real light field data. For the synthetic experiments, we compare our methods with five other state-of-art disparity estimation methods. From the experimental results, we conclude that our method can produce fairly accurate estimation results in scenes with different geometric structures and is robust to noise. When fine structures and occlusion exist in scenes, which often lead to severe mistakes for other disparity estimation methods, our method performs quite well. The state-of-the-art disparity estimation results are obtained without using guided filter and only using TV model optimization. With TV 
refinement, the estimation quality in smooth regions improves greatly, and the accuracy in occlusion regions can be maintained.

Compared with the deep-learning-based disparity estimation, the proposed matching entropy method is applicable for all kinds of light field data and is not affected by the acquisition method and scenario type, while the deep-learning-based method relies on the training data set and the application scenarios are limited. Especially for the light field data of the actual scene with the diversity of acquisition methods and the actual factors, the advantages of deep-learning-based methods cannot be realized. On the other hand, the deep-learning-based method usually integrates the data processing of the light field (such as super-resolution, and denoising) into the disparity estimation network, while the proposed matching entropy method directly processes the original light field data to generate the disparity estimation result.

 The disparity estimation performance in occluded regions was preliminarily verified. The real data experiment on the Tiger Piran scene and the simulation data experiment on the Pillow scene, show that the proposed method can reconstruct the disparity of the occlusion regions in complex situations with multiple layers of occlusion. However, the prerequisite for region type identification is segmentation. If the scene is too complicated to implement the segmentation, it will affect the reconstruction results of the occlusion regions. The limitation of the proposed matching entropy is that scene segmentation is an important premise for our adaptive region identification method. We will try to establish a region identification method without segmentation in future work.

\section{Conclusion}
To accurately estimate disparity information from light field data, we propose an adaptive region matching method to match sub-aperture images. Our main contributions are introducing the concept of matching entropy to measure the amount of correct matching information and designing a two-step adaptive process to select optimal matching windows in different regions. From the synthetic and real experiments, we verify that the proposed method can achieve high-precision disparity estimation of light field data, especially in occlusion and smooth regions, and is robust to noise. The core idea of defining matching entropy and selecting optimal matching windows adaptively is to treat regions differently according to their characteristics. This idea is not limited to light field data and can also be applied to area matching in more general stereo matching problems.
\bibliography{sample}
\end{document}